\documentclass[10pt,twocolumn,letterpaper]{article}

\usepackage{cvpr}              % To produce the CAMERA-READY version
\usepackage[dvipsnames]{xcolor}

\newcommand{\EZ}[1]{\iftrue {\color{orange}[EZ: #1]}\else {}\fi}

\definecolor{cvprblue}{rgb}{0.21,0.49,0.74}
\usepackage[pagebackref,breaklinks,colorlinks,citecolor=cvprblue]{hyperref}
\usepackage{multirow}

\newcommand{\cmmnt}[1]{\ignorespaces}
\DeclareMathOperator*{\argmax}{arg\,max} % Jan Hlavace
\def\paperID{8280} % *** Enter the Paper ID here
\def\confName{CVPR}
\def\confYear{2024}

\title{VOODOO 3D: \underline{Vo}lumetric P\underline{o}rtrait \underline{D}isentanglement f\underline{o}r \\ \underline{O}ne-Shot 3D Head Reenactment}

\author{
Phong Tran$^1$ \quad Egor Zakharov$^2$ \quad Long-Nhat Ho$^1$ \quad Anh Tuan Tran$^3$ \quad Liwen Hu$^4$\quad Hao Li$^{1,4}$\\ 
\small{\textsuperscript{1}MBZUAI \quad \textsuperscript{2}ETH Zurich \quad \textsuperscript{3}VinAI Research
\quad \textsuperscript{4}Pinscreen}\\
\texttt{\scriptsize \{the.tran, long.ho\}@mbzuai.ac.ae} \quad 
\texttt{\scriptsize anhtt152@vinai.io} \quad
\texttt{\scriptsize ezakharov@ethz.ch} \\
\texttt{\scriptsize liwen@pinscreen.com} \quad \texttt{\scriptsize  hao@hao-li.com}
}

\begin{document}
\twocolumn[{%
\renewcommand\twocolumn[1][]{#1}%
\maketitle
\begin{center}
    \centering
    \captionsetup{type=figure}
    \includegraphics[width=0.9\textwidth]{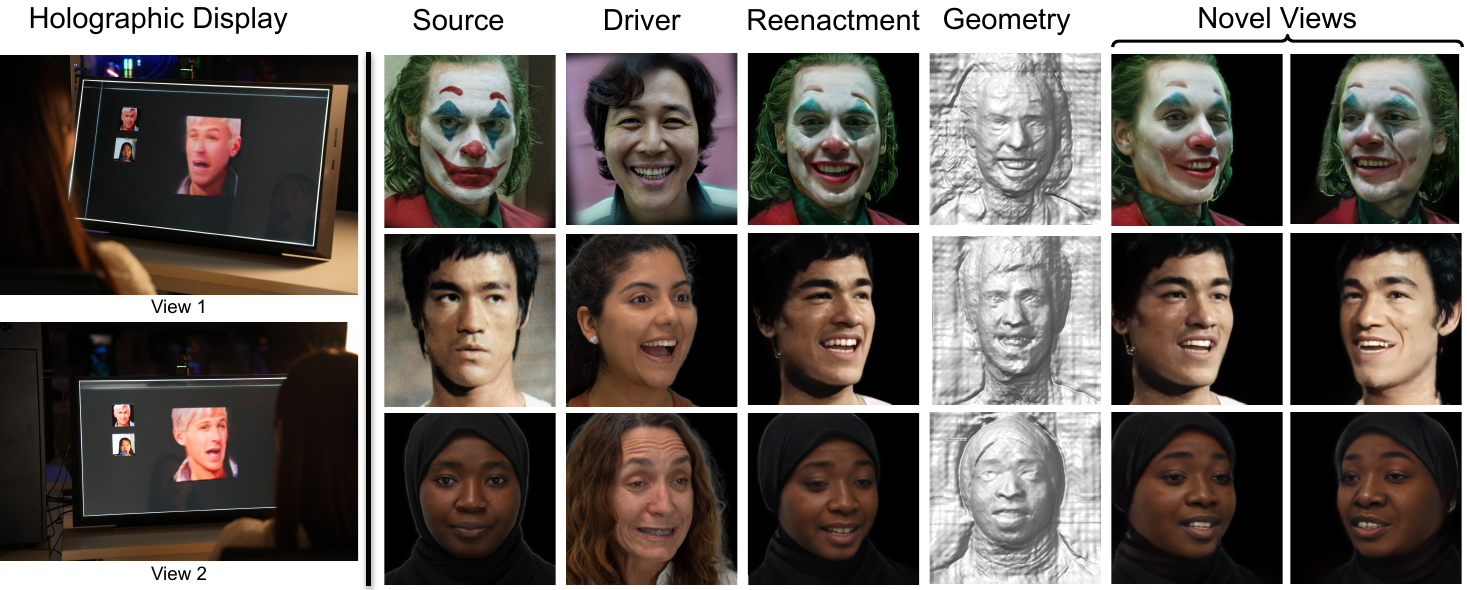}
    \parbox{14.5cm}{\captionof{figure}{We introduce \textbf{VOODOO 3D}: a high-fidelity 3D-aware one-shot head reenactment technique. Our method transfers the expression of a driver to a source and produces view consistent renderings for holographic displays.}}
    \label{fig:teaser}
\end{center}%
}]

\linespread{0.92}

\begin{abstract}

We present a 3D-aware one-shot head reenactment method based on a fully volumetric neural disentanglement framework for source appearance and driver expressions. 
Our method is real-time and produces high-fidelity and view-consistent output, suitable for 3D teleconferencing systems based on holographic displays. 
Existing cutting-edge 3D-aware reenactment methods often use neural radiance fields or 3D meshes to produce view-consistent appearance encoding, but, at the same time, they rely on linear face models, such as 3DMM, to achieve its disentanglement with facial expressions. 
%Cutting edge 3D aware methods often use triplane neural radiance fields or 3D meshes to enforce view consistent appearance encoding but typically rely on low dimensional linear face models (e.g., 3DMM, blendshapes, etc.) to extract facial expressions. 
%
As a result, their reenactment results often exhibit identity leakage from the driver or have unnatural expressions.
% As a result, the reenactments of these techniques often exhibit identity leak from the driver and unnatural expressions when rendered from angle that is too different than the source. 
%
To address these problems, we propose a neural self-supervised disentanglement approach that lifts both the source image and driver video frame into a shared 3D volumetric representation based on tri-planes.
% We propose a neural disentanglement approach that lifts both source image and driver video frame into canonical 3D volumetric representations based on triplanes. 
%
This representation can then be freely manipulated with expression tri-planes extracted from the driving images and rendered from an arbitrary view using neural radiance fields.
We achieve this disentanglement via self-supervised learning on a large in-the-wild video dataset.
%
% While this pose normalized triplane radiance field is used to encode 3D aware facial appearance features of the source, we also use it to frontalize the faces to extract high-fidelity appearance features for the source, as well as expressions for the driver. 
%
We further introduce a highly effective fine-tuning approach to improve the generalizability of the 3D lifting using the same real-world data.
%We further introduce a highly effective fine-tuning approach to the improve the generalizability of the 3D lifting using real world data. 
%
We demonstrate state-of-the-art performance on a wide range of datasets, and also showcase high-quality 3D-aware head reenactment on highly challenging and diverse subjects, including non-frontal head poses and complex expressions for both source and driver.

\end{abstract}    
\section{Introduction}
\label{sec:intro}

Creating 3D head avatars from a single photo is a core capability in making a wide range of consumer AR/VR and telepresence applications more accessible, and user experiences more engaging. 
%Creating 3D avatar heads from a single 2D photo is a core capability in making a range of consumer applications highly accessible and user experiences more engaging with personalized characters. 
%
Graphics engine-based 3D avatar digitization methods~\cite{10.1145/311535.311556,10.1145/3130800.31310887,DBLP:journals/corr/abs-2106-11423,gecer2021fast,lattas2021avatarme++,Lattas_2023_CVPR,Bai_2023_CVPR_FFHQ,Lei_2023_CVPR} are suitable for today's video games and virtual worlds, and many commercial solutions exist (AvatarNeo~\cite{pinscreenavatar}, AvatarSDK~\cite{avatarsdk}, ReadyPlayerMe~\cite{readyplayerme}, in3D~\cite{in3d}, etc.). However, the photorealism achieved by modern neural head reenactment techniques is becoming increasingly appealing for advanced effects in video sharing apps and visual effects. 
For immersive telepresence systems that use AR/VR headsets, facial expression capture is typically achieved using tiny video cameras built into HMDs~\cite{li2015facial,olszewski2016high,Lombardi_2018,Schwartz_2020_eye,Fu_2023_CVPR} \cmmnt{(Meta’s Avatar Codec)}, while the identity of the source subject recorded using a separate process.
However, the teleconferencing solutions based on holographic 3D displays (LookingGlass~\cite{lookinglass}, LEIA~\cite{leia}, etc.) use regular webcams \cite{trevithick2023real} \cmmnt{(NVIDIA’s LP3D)} or depth sensors \cite{Jason_2021_Starline}. \cmmnt{(Google’s Project Starline)} 
As opposed to a video-based setting, head reenactment for immersive applications needs to be 3D-aware, meaning that in addition to generating the correct poses and expressions from a photo, multi-view consistency is critical.

While impressive facial reenactments results have been demonstrated using 2D approaches ~\cite{zakharov2020fast,doukas2020headgan,wang2021one,StyleHEAT_2022,drobyshev2022megaportraits,Zhang_2023_CVPR}, \cmmnt{(e.g., MegaPortrait, etc.),} they typically struggle with preserving the likeness of the source and exhibit significant identity changes when varying the camera pose. 
More recently, 3D-aware one-shot head reenactment methods ~\cite{Khakhulin2022ROME,hong2022headnerf,ma2023otavatar,Li_2023_CVPR,li2023generalizable,yu_2023_nofa} \cmmnt{(LPR, NOFA, ROME)} have used either 3D meshes or tri-plane neural radiance fields as a fast and memory efficient volumetric data representations for neural rendering. 
However, the expression and identity disentanglement in these methods is based on variants of linear face and expression models \cite{blanz1999morphable,li2017learning} \cmmnt{(3DMM, FLAME, etc.)} which lack expressiveness and high-frequency details. 
While these methods can achieve view consistency, facial expressions are often uncanny, and preserving the likeness of the input source portrait is challenging, especially for views different than the source image. Hence, input sources with extreme expressions and non-frontal poses are often avoided.% since facial identity and expression disentanglement can be problematic. 

In this paper, we introduce the first 3D aware one-shot head reenactment technique that disentangles source identities and the target expressions fully volumetrically, and without the use of explicit linear face models. 
Our method is real-time and designed with holographic displays in mind, where a large number of views (up to 45) can be rendered in parallel based on their viewing angle. 
We leverage the fact that real-time 3D lifting for human heads has recently been made possible~\cite{trevithick2023real} with the help of Vision Transformers (ViT)~\cite{Dosovitskiy2020AnII}, which avoids the need for inefficient optimization-based GAN-inversion process~\cite{roich2021pivotal}.
In particular, 3D lifting allows us to map 2D face images into a canonical tri-plane representation for both source and target subjects and treat identity and expression disentanglement independently from the head pose. 

Once the source image and driver frame are lifted into a pose-normalized tri-plane representation, we extract appearance features from the source subject and expressions from the driver. 
The pose of the driver is estimated separately using a 3D face tracker and used as input to a neural renderer.
Tri-plane-based feature extraction ensures view-consistent rendering, while facial appearance and driver expression feature use frontalized views from the 3D lifting to enable robust and high-fidelity facial disentanglement.
To handle highly diverse portraits (variations in facial appearance, hairstyle, head covering, eyewear, etc.), we propose a new method for fine-tuning Lp3D on real datasets by introducing a mixed loss function based on real and synthetic datasets. 
Our volumetric disentanglement and rendering framework is trained only using in-the-wild videos from the CelebV-HQ dataset~\cite{zhu2022celebvhq} in a self-supervised fashion.

We not only demonstrate that our volumetric face disentanglement approach produces qualitative superior head reenactments than existing ones, but also show on a wide and diverse set of source images how non-frontal poses and extreme expressions can be handled. 
We have quantitatively assessed our method on multiple benchmarks and outperform existing 2D and 3D state-of-the-art techniques in terms of fidelity, expression, and likeness accuracy metrics. 
Our 3D aware head reenactment technique is therefore suitable for AR/VR-based immersive applications, and we also showcase a teleconferencing system using a holographic display from LookingGlass~\cite{lookinglass}.
We summarize the main contributions as follows:
\begin{itemize}
    \item First fully volumetric disentanglement approach for real-time 3D aware head reenactment from a single photo. This method combines 3D lifting into a canonical tri-plane representation and formalized facial appearance and expression feature extraction.
    \item A 3D lifting network that is fine-tuned on unconstrained real-world data instead of only generating synthetic ones.
    \item We demonstrate superior fidelity, identity preservation, and robustness w.r.t. current state-of-the-art methods for facial reenactment on a wide range of public datasets. We plan to release our code to the public.
    %\item We also showcase an avatar-based reenactment system for immersive communication using a holographic display, and we plan to release our code to the public.
\end{itemize}

\section{Related Work}

%TODO: Describe video-based and multiview-based avatars that require lengthy fine-tuning. Describe single-shot methods such as Megaportraits and mention that they do not produce view-consistent results. At the same time, 3D avatars ROME and LPR achieve that. We are better because we do not rely on 3DMM parameters and thus achieve higher fidelity.

\paragraph{2D Neural Head Reenactment.} 
The problem of generating animations of photorealistic human heads given images or video inputs has been thoroughly explored using various neural rendering techniques in the past few years, outperforming traditional 3DMM-based methods~\cite{thies2016face,thies2018headon,kim2018deepvideo,doukas2020headgan,gafni2021nerface,yangface2face,nirkin2019fsgan,Ren_2021_ICCV,Athar_2022_CVPR} which often appear uncanny due to their compressed linear space.
%been thoroughly explored in the literature~\cite{TODO}.
These approaches can be categorized into one-shot
%~\cite{TODO} 
and multi-shot ones.
%~\cite{TODO}.
While multi-shot methods generally achieve high-fidelity results, they are not suitable for many consumer applications as they typically require an extensive amount of training data, such as a monocular video capture~\cite{gafni2021dynamic,Cao2022Authentic,Zheng_2022_CVPR,Grassal_2022_CVPR,Zheng2023pointavatar,bharadwaj2023flare,Xu2023LatentAvatar,INSTA2023Zielonka,Chen_2023_CVPR,HAvatar2023Zhao,Bai_2023_CVPR}, and sometimes even a calibrated multi-view stereo setup~\cite{Lombardi_2018,Schwartz_2020_eye,Ma_2021_CVPR,Bi2021DeepRA,Fu_2023_CVPR}. %, which is typically associated with hours or even days of training to obtain a single avatar.
More recently, few-shot techniques~\cite{zhang2022fdnerf} have also been introduced.
%\EZ{can add much more works here, lets do that later} and hours or even days of training to obtain each avatar.
%Thus, these methods remain impractical for the majority of applications that require human avatars.

%In parallel, a considerable subset of methods
To maximize accessibility, a considerable number of methods
~\cite{Wiles18,Siarohin_2019_CVPR,Siarohin_2019_NeurIPS,zakharov2019few,burkov2020neural,zakharov2020fast,doukas2020headgan,song2021pareidolia,wang2021latent,wang2021one,Ren_2021_ICCV,siarohin2021motion,EAMM_2022,hong2022depth,Tao_2022_CVPR,StyleHEAT_2022,drobyshev2022megaportraits,Zhao_2022_CVPR,Zhang_2023_CVPR,Gao_2023_CVPR} use a single portrait as input by leveraging advanced generative modeling techniques based on in-the-wild video training data. While most methods rely on linear face models to extract facial expressions, the head reenactment technique from Drobyshev et al.~\cite{drobyshev2022megaportraits} directly extract expression features from cropped 2D face regions, allowing them to obtain better face disentanglements, which results in higher fidelity face synthesis. While similar to our proposed approach in avoiding the use of low dimensional linear face models, their method is purely 2D and struggly with ensuring identity and expression consistency when novel views are synthesized.

%
%However, The majority of these approaches~\cite{zakharov2020fast,doukas2020headgan,wang2021one,StyleHEAT_2022,drobyshev2022megaportraits,Khakhulin2022ROME} relied purely on modeling 2D image-based data, resulting in poor identity and expression consistency for novel view synthesis.
%
\paragraph{3D-Aware One-Shot Head Reenactment.} 
Due to potential inconsistencies when rendering from different views or poses, a number of 3D-aware single shot head reenactment techniques~\cite{Schwarz2020graf,Chan_2021_CVPR,Niemeyer_2021_CVPR,chan2022efficient,Or-El_2022_CVPR,epigraf2022,Xue_2022_CVPR,Deng_2022_CVPR,Xiang_2023_ICCV,An_2023_CVPR,xu2022pv3d,Xu_2023_CVPR} have been introduced. These methods generally use an efficient 3D representation, such as neural radiance fields or 3D mesh, to geometrically constraint the neural rendering and improve view consistency. ROME~\cite{Khakhulin2022ROME} for instance is a mesh-based method using FLAME blendshapes~\cite{FLAME:SiggraphAsia2017} and neural textures. While view-consistent results can be produced for both face and hair regions, the use of low resolution polygonal meshes hinders the neural renderer to generate high-fidelity geometric and appearance details. 

Implicit representations such as HeadNeRF~\cite{hong2022headnerf} and MofaNeRF~\cite{Zhuang_2022_ECCV} use a NeRF-based parametric model which supports direct control of the head pose of the generated images. While real-time rendering is possible, these methods require intensive test-time optimization and often fail to preserve the identity of the source due to the use of compact latent vectors. Most recent methods~\cite{Li_2023_CVPR,yu_2023_nofa,li2023generalizable} adopt the highly efficient tri-plane-based neural fields representation~\cite{chan2022efficient} to encode the 3D structure and appearance of the avatars head. Compared to the previous works on view-consistent neural avatars~\cite{Khakhulin2022ROME,hong2022headnerf,ma2023otavatar,Li_2023_CVPR,li2023generalizable,yu_2023_nofa}, we refrain from depending on parametric head models for motion synthesis and, instead, learn the volumetric motion model from the training data.
This methodology enables us to narrow the identity gap between the source and generated images and yield a superior fidelity of the generated motion compared to competing approaches, and hence a higher quality disentanglement for reenactment.

%Li et al. 2023\cite{Li_2023_CVPR}

%NOFA~\cite{yu_2023_nofa} 

%LPR~\cite{li2023generalizable}

%Nevertheless, the recent development of 3D-aware generative models, which are based on accurate geometric reconstruction of the animated subject and its subsequent rendering, hold the potential of realizing view-consistent single-shot avatars.
%
%Our work follows these 3D-aware generative models and employs the triplane representation~\cite{chan2022efficient} to encode the 3D structure and appearance of the avatar.
%

%TODO: Describe 3D GANs such as EG3D and their inversion approaches: optimization-based inversions such as PTI, encoder-based inversions such as Triplane-Net, and LP3D. Encoder-based inversion offers favorable speed but does not disentangle appearance and expression, which we introduce in our work.

\paragraph{3D GAN Inversion.} 
%A highly relevant body of work to ours is 3D GAN inversion.
%Our work is also closely linked to the problem of 3D GAN inversion.
%
When training a whole reconstruction and disentangled reenactment model end-to-end on facial performance videos, one can introduce substantial overfitting and reduce the quality of the results.
To address this problems, we focus our training approach to an inversion of pre-trained 3D-aware generative models for human heads.

We use tri-plane-based generative network EG3D~\cite{chan2022efficient} as the foundational generator, due to its proficiency in producing high-fidelity and view-consistent synthesis of human heads.
%Its 3D GAN inversion methods~\cite{roich2021pivotal} leverage these properties to estimate latent representations for a given image that are decoded into the view-consistent outputs and match the contents of the input.
%
For a given image, an effective 3D GAN inversion method should leverage these properties for estimating latent representations, which can be decoded into outputs that maintain view consistency and faithfully replicate the contents of the input.
One naive approach is to adapt GAN inversion methods that were initially designed for 2D GANs to the EG3D pre-trained network. 
These methods either do a time consuming but more precise optimization \cite{Karras_2020_CVPR,roich2021pivotal} or train a fast but less accurate encoder network \cite{richardson2021encoding,tov2021designing} to obtain the corresponding latent vectors. 
They often produce incorrect depth prediction, leading to clear artifacts in novel view synthesis. 
Hence, some methods are specifically designed for inverting 3D GANs, which either do multi-view optimization \cite{ko20233d,xie2023high} or predict residual features/tri-plane maps for refining the initial inversion results \cite{yuan2023make,bhattarai2023triplanenet,Yin_2023_CVPR, trevithick2023real}.%, which are cumbersome. 
%
%Instead of predicting latents in the common compact spaces like $\mathcal{W}$+ or $\mathcal{S}$+, the state-of-the-art EG3D inversion method called Lp3D~\cite{trevithick2023real} achieves real-time performance and high accuracy of novel-view reenactment by directly predicting triplanes from the input image via a visual transformer (ViT)~\cite{xie2021segformer}.
%Among the more popular inversion methods are optmization-based~\cite{TODO}, such as PTI~\cite{roich2021pivotal}, which solve an optimization problem to obtain the corresponding latent vectors.
%These methods are usually extremely slow and require minutes to invert each frame, thus making them highly impractical.
%Another group of inversion methods~\cite{TODO} achieves more favorable runtime characteristics by directly predicting the latent vectors given an input image via a pre-trained encoder network.

In this work, we rely on the state-of-the-art EG3D inversion method Lp3D~\cite{trevithick2023real}.
While achieving excellent novel-view synthesis results, it lacks disentanglement between the appearance and expression of the provided image and is unable to impose various driving expressions onto the input.
To address this limitation, we propose a new method that introduces appearance-expression disentanglement in the latent space of tri-planes using our new self and cross-reenactment training pipeline while relying on a pre-trained but fine-tuned Lp3D network for regularization which enables highly consistent view synthesis.

\section{3D-Aware Head Reenactment}

%\subsection{Overview}

\begin{figure*}[t]
    \centering
    \includegraphics[width=0.85\textwidth]{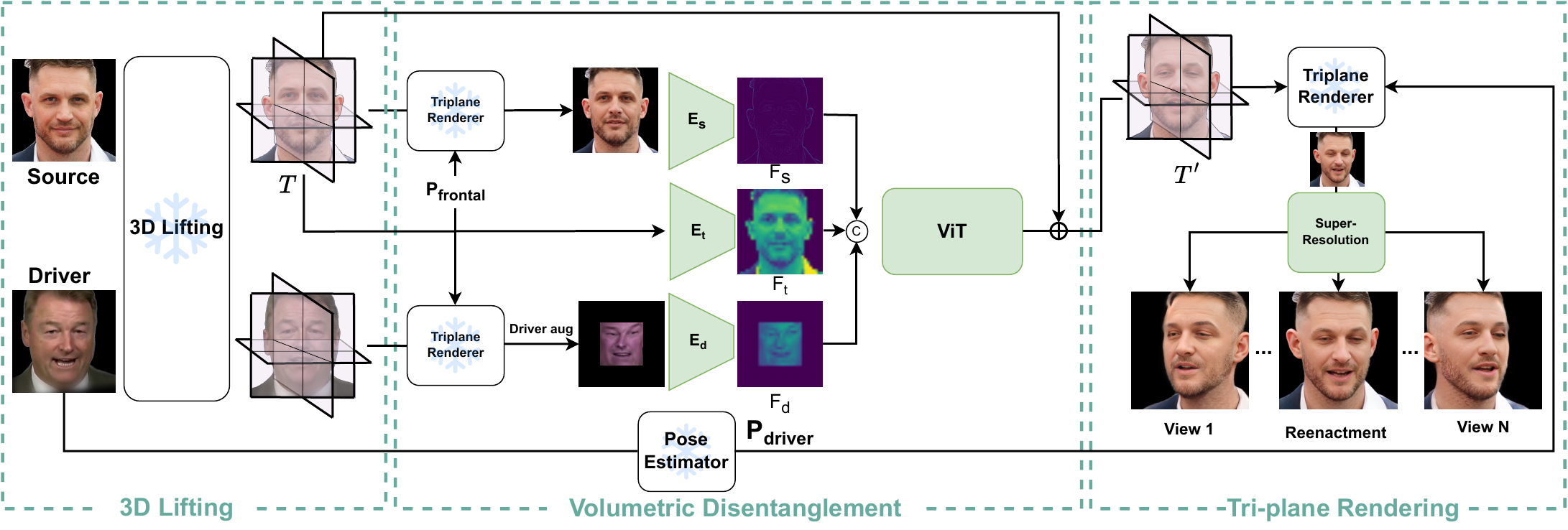}
    \caption{%Given a pair of source and driver images, we first frontalize them using a pretrained triplane-based 3D lifting module \cite{trevithick2023real}. Then, the frontalized images are fed into two separate convolutional encoders to extract the face features $F_s$ and $F_d$. These extracted features are concatenated with the features extracted from the triplane of the source, and all are fed together into a transformer block \cite{xie2021segformer} to produce the expression triplane residual, which is added to the triplane of the source image. The final target image can be rendered from the new triplane using a pre-trained triplane renderer with the driver's pose.\anh{Should the triplane encoder be inside the VOODOO block since it is finetuned by a novel mechanism? Increase text size and line thickness}
    Given a pair of source and driver images, our method processes them in three steps: \textbf{3D Lifting} into tri-plane representations, \textbf{Volumetric Disentanglement}, which consists of source and driver frontalization and tri-plane residual generation, and \textbf{Tri-plane Rendering} via volumetric ray marching with subsequent super-resolution.
    }
    \label{fig:arch}
\end{figure*}

As illustrated in Fig.~\ref{fig:arch}, our head reenactment pipeline consists of three stages: 1) 3D Lifting, 2) Volumetric Disentanglement, and 3) Tri-plane Rendering.
Given a pair of source and driver images, we first frontalize them using a pre-trained but fine-tuned tri-plane-based 3D lifting module \cite{trevithick2023real}.
This driver alignment step is crucial and allows our model to disentangle the expressions from the head pose, which prevents overfitting.
Then, the frontalized faces are fed into two separate convolutional encoders to extract the face features $F_s$ and $F_d$. 
These extracted features are concatenated with the ones extracted from the tri-planes of the source, and all are fed together into several transformer blocks~\cite{xie2021segformer} to produce the expression tri-plane residual, which is added to the tri-planes of the source image. 
The final target image can be rendered from the new tri-planes using a pre-trained tri-plane renderer using the driver's pose.

%\anh{Add paragraph(s) for problem definition and system overview. Introduce the system components (3D lifting module, Expression module) and why we need them.}

\subsection{Fine-Tuned 3D Lifting}
We adopt Lp3d \cite{trevithick2023real} as a 3D face-lifting module, which predicts the radiance field of any given face image in real-time. Instead of using an implicit multi-layer perceptron \cite{mildenhall2021nerf} or sparse voxels \cite{fridovich2022plenoxels,schwarz2022voxgraf} for the radiance field, Lp3D \cite{trevithick2023real} uses tri-planes~\cite{chan2022efficient}, which can be computed using a single forward of a deep learning network. Specifically, for a given source image $x_s$, we first extract the tri-planes $T$ using a transformer-based appearance encoder $\mathbf{E}_{\text{app}}$: %\anh{denotations for network modules and data currently have the same format, making the paper uneasy to read. May use different format to differentiate them, e.g., use mathbf for network module denotations}:
\begin{align}
    \mathbf{E}_{\text{app}}(x_s) = T \in \mathbb{R}^{3 \times H \times W \times C} = \{T_{xy}, T_{yz}, T_{zx}\}.
\end{align}
The color $c$ and density $\sigma$ of each point $p = (x, y, z)$ in the radiance field can be obtained by projecting $p$ onto the three planes and by summing up the features at the projected positions:
\begin{align}
    c, \sigma = \mathbf{D}(F_{xy} + F_{yz} +F_{zx}),
\end{align}
where $\mathbf{D}$ is a shallow MLP decoder for the tri-plane rendering, $F_{xy}, F_{yz}$, and $F_{zx}$ are the feature vectors at the projected positions on $xy, yz$, and $zx$ planes, respectively, calculated using bilinear interpolation. The rendered $128 \times 128$ image is then upsampled using a super-resolution module to produce a high-resolution output. To train the encoder $\mathbf{E}_{\text{app}}$, Lp3D \cite{trevithick2023real} uses synthetic data generated from a 3D-aware face generative model \cite{chan2022efficient}. While these synthetic data have ground truth camera poses, they are limited to the face distribution of the generative model. As a result, Lp3D can fail to generalize to in-the-wild images as shown in~\cref{fig:lp3dfinetune}. To prevent this, we fine-tune the pre-trained Lp3D on a large-scale real-world dataset. We also replace the original super-resolution module in Lp3D with a pre-trained GFPGAN \cite{wang2021towards}, which is then fine-tuned together with Lp3D (see~\cref{sec:training}).

\subsection{Disentangling Appearance and Expression}
%Controlling facial expression in a 3D radiance field is very challenging. %A good expression module must be: (1) light-weight; (2) able to fully disentangle expressions and identity; (3) able to corresponde expressions between unaligned source and driver images; and (4) able to produce realistic expressions \anh{[these conditions sound a bit random and unorganized}. One way to achieve (1) and (2) is to use an off-the-shelf morphable-based face model \anh{do you mean the 3DMM model or the modeling method?} \cite{deng2019accurate} that can fully disentangle the face shape and expression. For example, LPR \cite{li2023generalizable} predicts 3DMM face coefficients \cite{blanz1999morphable} including the shape and expression of both source and driver, then calculates the canonical source's mesh with driver's expression using these coefficients. This mesh is then used to modify the expression of the source's triplane to the driver. While being able to disentangle the expression fully, this approach is highly dependent on the 3DMM face model, which often has inaccurate and robot-like expressions.
%To reenact the source image using the driver's facial expressions in a 3D radience field, we need to separate 

%We build an expression module on top of the Lp3D model that satisfies all the aforementioned criteria. The module's architecture is given in \cref{fig:arch}. 

Separating facial expression from the identity appearance in a 3D radiance field is very challenging especially when source and driver subjects have misaligned expressions. In order to simplify the problem, we use our 3D lifting approach to bring both source and driver heads into a pose-oriented space where faces are frontalized. Here, we denote frontalized source and driver images as $x_s^f$ and $x_d^f$, respectively. These images are then fed into two separate convolutional source and driver encoders $\mathbf{E}_s$ and $\mathbf{E}_d$ to produce coarse feature maps:
 \begin{align*}
     F_s &= \mathbf{E}_s(x_s^f)\\
     F_d &= \mathbf{E}_d(x_d^f)
 \end{align*}
Since we already have the source's tri-plane, which encodes the 3D shape of the source, we use another encoder to encode this tri-plane and concatenate it together with the coarse frontalized feature maps of the images to produce expression feature $F$:
 \begin{align*}
     F_t &= \mathbf{E}_t(T)\\
     F &= F_s \oplus F_d \oplus F_t
 \end{align*}
 Even though face frontalization aligns the source and the driver, there is still some misalignment between the two faces, e.g., the positions of the eyes may be different, or one mouth is open while the other is closed. Therefore, we feed the concatenation of the feature maps into several transformer blocks to produce the final residual tri-plane $\mathbf{E}_v(F)$. This residual is then added back to the source's tri-planes to change the source's expression to the driver's expression $T' = T + \mathbf{E}_v(F)$. Unlike LPR \cite{li2023generalizable}, we do not use a 3D face model to compute the expression but instead use the RGB images of the source and the driver directly, allowing the model to learn high-fidelity and realistic expressions.

\subsection{Tri-Plane Rendering}

%explain triplane rendering and super resolution module, and how we render multiple views in parallel for holographic displays in addition to the current estimated pose. Also mention which pose estimator we use.

The resulting tri-planes are then volumetrically rendered into one or multiple output images using pose parameters and viewing angles in the case of a holographic display.
Following EG3D~\cite{chan2022efficient}, we use a neural radiance fields (NeRFs)-based volumetric ray marching approach~\cite{mildenhall2020nerf}.
However, instead of encoding each point in space via positional encodings~\cite{mildenhall2020nerf}, the features of the points along rays are calculated using their projections onto tri-planes.
%
%Therefore, triplanes can be viewed as a factorized volumetric feature grid, which is processed by a volume rendering multi-layer perception (MLP) to produce point-wise color and density.
%
Since tri-planes are aligned with the frontal face, we can compute these rays directly using camera extrinsics $\text{P}_{\text{driver}}$ predicted by an off-the-shelf 3D head pose estimator~\cite{deng2019accurate}.

While the renderings are highly view-consistent, the large number of points evaluated for each ray still limits the ouput resolution for real-time performance. We therefore follow~\cite{li2023generalizable} and employ a 2D upsampling network~\cite{wang2021gfpgan} based on StyleGAN2~\cite{Karras_2019_CVPR}, which in our experiments produced higher quality results than the upsampling approach in EG3D~\cite{chan2022efficient}.
Finally, for holographic displays, we generate a number of renderings based on their viewing angles and simply using the head pose parameter.
Real-time performance is achieved using efficient inference libraries such as TensorRT, half-precision, and batched inference over multiple GPUs.

\subsection{Training Strategy} \label{sec:training}
\paragraph{Fine-Tuning Lp3D.} To make Lp3D work with in-the-wild images, we fine-tune it on a large-scale real-world video dataset \cite{zhu2022celebv}. Unlike the use of synthetic data, real-world data do not have ground-truth camera parameters and facial expressions in monocular videos are typically inconsistent over time. While the camera parameters can be estimated using standard 3D pose estimators, the expression diferences are difficult to determine. However, we found that we can ignore this expression difference and fine-tune Lp3D using real data together with continuous training on synthetic data. In particular, our experiments indicate that the fine-tuned model can still faithfully reconstruct 3D faces from the input without changing expressions and still generalize successfully on in-the-wild images. 
%\anh{weird, may need to provide evidence, at least in the supplementary}. 
Specifically, on real video data, we sample two frames $x_s^r$ and $x_d^r$ and estimate their camera paramters $P_s^r$ and $P_d^r$. Similar to~\cite{chan2022efficient}, we assume a fixed intrinsics for standard portraits for all images. Then we use $E_{app}$ from Lp3D to calculate the tri-planes of $x_s^r$, render it using the two poses, and calculate reconstruction losses on the two rendered images:
\begin{align*}
    \mathcal{L}_{\text{real}} = \|\text{Lp3D}(x_s^r, P_d^r) - x_d^r\| + \|\text{Lp3D}(x_s^r, x_s^r) - x_s^r\|,
\end{align*}
where $\text{Lp3D}(x, P)$ is the face in $x$ re-rendered using camera pose $P$ and $\mathcal{L}_{\text{real}}$ is the loss for real images. Simultaneously, we render two synthetic images employing an identical latent code but through varying camera views and calculate the synthetic loss $\mathcal{L}_{\text{syn}}$:
\begin{align*}
    \mathcal{L}_{\text{syn}} &= \|\text{Lp3D}(x_s^f, P_d^s) - x_d^s\| + \|\text{Lp3D}(x_s^s, P_s^s) - x_s^s\| \\
    \mathcal{L}_{\text{tri}} &= \|E_{\text{app}}(x_s^f) - T\|,
\end{align*}
where $T$ is the ground-truth tri-planes returned by EG3D \cite{chan2022efficient} and $\mathcal{L}_{\text{tri}}$ is the tri-plane loss adopted directly from Lp3D. The final loss $\mathcal{L}_{\text{app}}$ for fine-tuning Lp3D can be formulated as: 
\begin{align*}
    \mathcal{L}_{\text{app}} = \mathcal{L}_{\text{real}} + \lambda_{\text{syn}} \mathcal{L}_{\text{syn}} + \lambda_{\text{tri}}\mathcal{L}_{\text{tri}}
\end{align*}
where $\lambda_{\text{syn}}$ and $\lambda_{\text{tri}}$ are tunable hyperparameters.

\paragraph{Disentangling Appearance and Expressions.}
In this stage, we also use real-world videos as training data. For a pair of source and driver images $x_s$ and $x_d$ sampled from the same video, we apply the reconstruction loss $\mathcal{L}_{\text{recon}}$ which is a combination of $L1$, perceptual \cite{zhang2018unreasonable}, and identity losses, between the reenacted image $x_{s \rightarrow d}$ and the corresponding ground-truth $x_d$:
\begin{align*}
    \mathcal{L}_{\text{recon}} &= \|x_{s \rightarrow d} - x_d\|_1 + \phi\left(x_{s \rightarrow d}, x_d\right) \\
    &+ \|\text{ID}(x_{s \rightarrow d}) - \text{ID}(x_d)\|_1,
\end{align*}
% \anh{explain notations $\phi$ and ID. Missing one loss term.}
where $\phi$ is the perceptual loss and ID($\cdot$) is a pretrained face recognition model. Similar to other works that use RGB images directly to calculate expressions \cite{drobyshev2022megaportraits}, our proposed encoder also suffers from an ``identity leaking" issue. Since there is no cross-reenactment dataset, the expression module is trained with self-reenactment video data. Therefore, without proper augmentation and regularization, the expression module can leak identity information from the driver to the output, making the model fail to generalize to cross-reenactment tasks. Hence, we introduce a \textit{Cross Identity Regularization}. Specifically, we further sample an additional driver frame $x_{d'}$ from another video. We incorporate a GAN loss where real samples are $\text{Lp3D}(x_s, P^d)$ and fake samples are $x_{s \rightarrow d'}$. This GAN loss is also conditioned on the identity vector of the source $\text{ID}(x_s)$. Following \cite{drobyshev2022megaportraits}, we also apply strong augmentation (random warping and color jittering) and additionally mask the border of the driver randomly to further reduce potential identity leaks. The loss for expression training can be summarized as:
\begin{align*}
    \mathcal{L}_{\text{exp}} = \mathcal{L}_{\text{recon}} + \lambda_{\text{CIR}} \mathcal{L}_{\text{CIR}},
\end{align*}
where $\mathcal{L}_{\text{CIR}}$ and $\lambda_{\text{CIR}}$ are cross identity regularization and its hyperparameter, respectively. 
% \liwen{the formula of $\mathcal{L}_{\text{CIR}}$ is not clear here}

\paragraph{Global Fine-Tuning.} After training both Lp3D and the expression module, we iteratively fine-tune the two modules using the same losses as the previous sections. Specifically, for every 10000 iterations, we freeze one module and fine-tune the other and vice versa. In addition, we add a GAN loss on the super-resolution output of the Lp3D module.

\begin{figure*}
    \centering
    \includegraphics[width=0.93\linewidth]{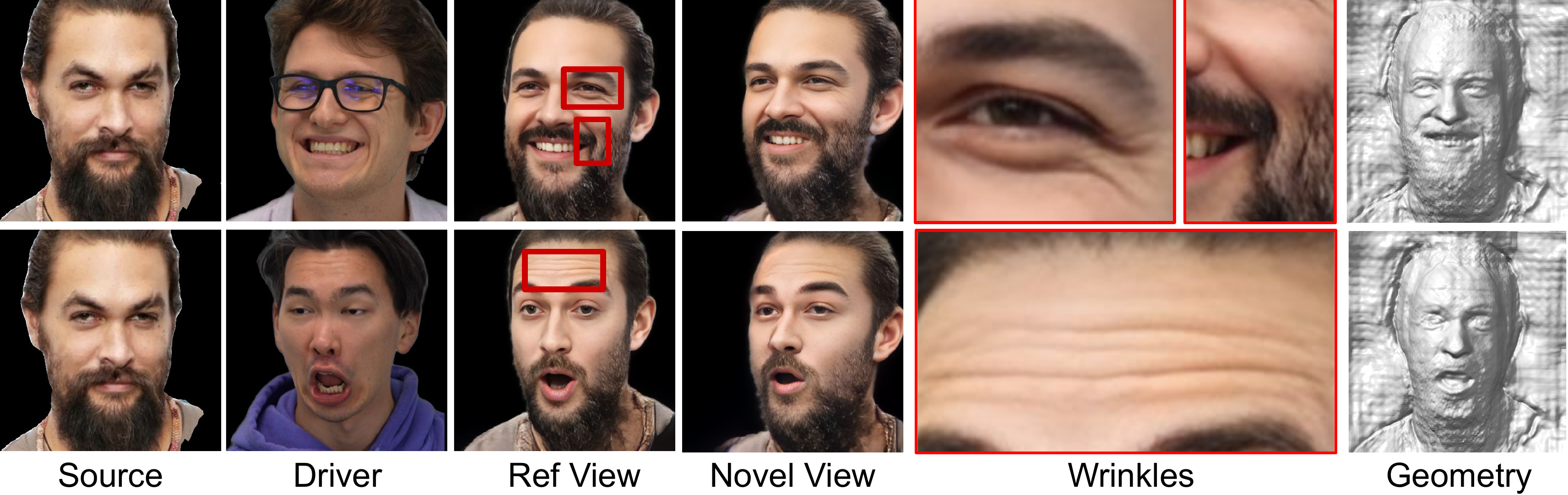}
    \vspace{-0.15cm}
    \caption{Expression dependent high-fidelity details, incl. eye and forehead wrinkles, as well as nasolabial folds (see zoom-ins)}
    \label{fig:wrinkles}
\end{figure*}

\begin{figure*}
    \centering
    \includegraphics[width=0.93\linewidth]{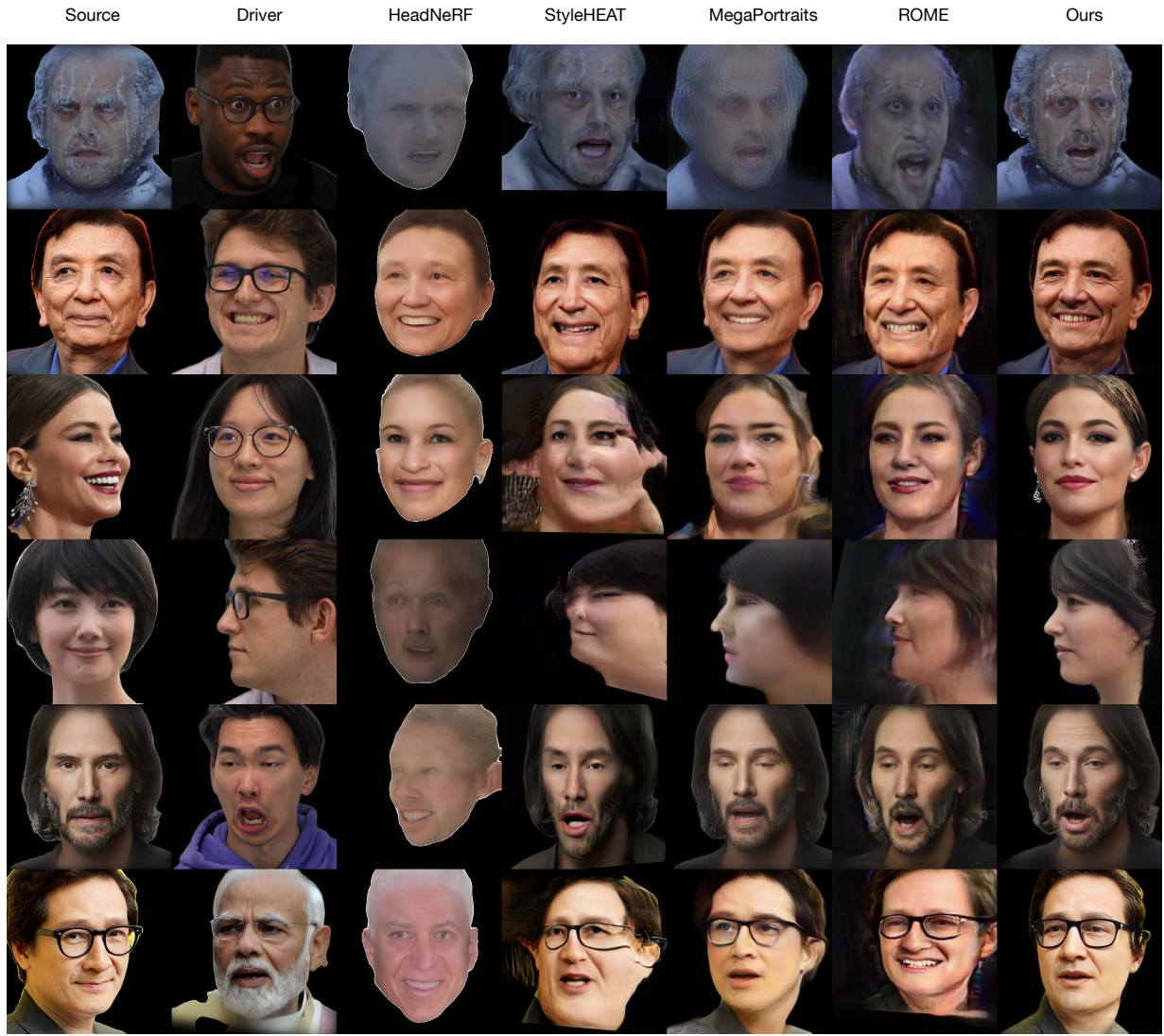}
    \vspace{-0.1cm}
    \caption{A qualitative comparison with the baselines on in-the-wild photos. Notice that our method is capable of producing a variety of facial expressions, and handle highly diverse subjects, with and without accessories, as well as extreme head poses, such as rows 3 and 4.}
    \label{fig:qual}
\end{figure*}
\section{Experiments}
\begin{figure*}[ht]
\begin{minipage}[t]{0.67\textwidth}
    \centering
    \vskip 0.05in
    \resizebox{\linewidth}{!}{%
    \begin{tabular}{l|cccccc|ccc}
        %\toprule
        \multirow{2}{*}{Method} & \multicolumn{6}{c|}{Self-reenactment} & \multicolumn{3}{c}{Cross-reenactment}\\
        & PSNR $\uparrow$ & SSIM $\uparrow$ & LPIPS $\downarrow$ & NAKD $\downarrow$ & ECMD $\downarrow$ & FID $\downarrow$ & CSIM $\uparrow$ & ECMD $\downarrow$ & FID $\downarrow$\\
        \hline
        ROME \cite{Khakhulin2022ROME} & 18.46 & 0.488 & 0.351 & 0.030 & 0.594 & 138 & 0.507 & \textbf{0.740} & 172 \\
        StyleHeat \cite{StyleHEAT_2022} & 19.73 & 0.689 & 0.278 & 0.035 & 0.748 & 89.8 & 0.398 & 0.744 & 95.5 \\
        OTAvatar \cite{ma2023otavatar} & 19.28 & 0.749 & 0.289 & 0.035 & 0.651 & 67.0 & 0.462 & 0.901 & 72.4 \\
        MegaPortraits \cite{drobyshev2022megaportraits} & 21.10 & 0.731 & 0.291 & 0.022 & 0.755 & 52.0 & 0.729 & 0.771 & 61.7\\
        Ours & \textbf{22.83} & \textbf{0.768} & \textbf{0.168} & \textbf{0.012} & \textbf{0.426} & \textbf{40.5} & \textbf{0.754} & 0.754 & \textbf{36.4}\\
         % \cmidrule(lr){2-4} \cmidrule(lr){2-4}
         %\bottomrule
    \end{tabular}
    }
    \captionof{table}{Evaluation on HDTF~\cite{zhang2021flow} dataset. Our method outperforms the competitors across almost all of the metrics for both self- and cross-reenactment scenarios.}
    \label{tab:hdtf}
    \vspace{-3mm}
\end{minipage}
\hfill
\begin{minipage}[t]{0.3\textwidth}
    \centering
    \vskip 0.05in
    \resizebox{\linewidth}{!}{%
    \begin{tabular}{l|ccc}
        %\toprule
        \multirow{2.5}{*}{Method} & \multicolumn{3}{c}{Cross-reenactment}\\
         & CSIM $\uparrow$ & ECMD $\downarrow$ & FID $\downarrow$\\
        \hline
        ROME \cite{Khakhulin2022ROME} & 0.519 & 0.91 & 52.6\\
        HeadNeRF \cite{hong2022headnerf} & 0.346 & 0.88 & 113 \\
        StyleHeat \cite{StyleHEAT_2022} & 0.467 & 0.85 & 50.2 \\
        MegaPortraits \cite{drobyshev2022megaportraits} & \textbf{0.647} & \textbf{0.77} & 29.2\\
        Ours & 0.608 & 0.79 & \textbf{23.6} \\
         % \cmidrule(lr){2-4} \cmidrule(lr){2-4}
         %\bottomrule
    \end{tabular}
    }
    \captionof{table}{Evaluation on CelebA-HQ~\cite{karras2017progressive} dataset.}
    \label{tab:celeba}
    \vspace{-3mm}
\end{minipage}
\end{figure*}

\begin{figure}[ht]
\begin{minipage}[t]{0.225\textwidth}
    \centering
    \vskip 0.05in
    \includegraphics[width=\linewidth]{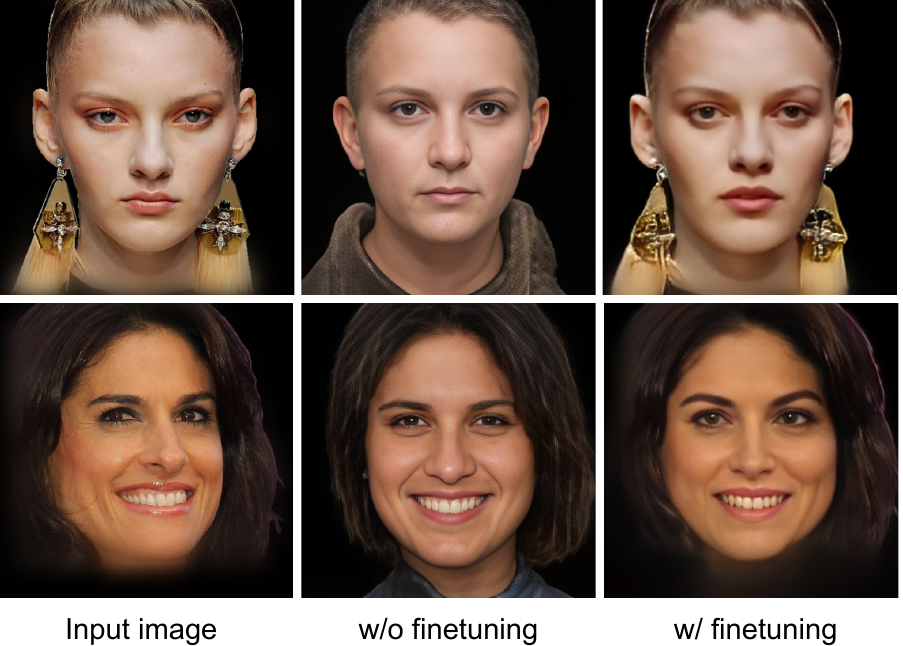}
    \vspace{-0.58cm}
    \captionof{figure}{Our implementation of Lp3D~\cite{trevithick2023real} before and after CelebV-HQ~\cite{zhu2022celebvhq} fine-tuning.}
    \label{fig:lp3dfinetune}
\end{minipage}
\hfill
\begin{minipage}[t]{0.225\textwidth}
    \centering
    \vskip 0.05in
    \resizebox{\linewidth}{!}{%
        \begin{tabular}{l|cc}
             & CSIM $\uparrow$ & ECMD $\downarrow$\\
            \hline
            Lp3D & 0.548 & 0.82 \\
            Lp3D-FT & 0.670 & 0.76\\
            w/o frontal & 0.668 & 1.01\\
            w/o CIR & 0.570 & 0.97 \\
            Ours & 0.608 & 0.79 \\
        \end{tabular}
        }
    \vspace{-0.25cm}
    \captionof{table}{Ablation studies conducted on CelebA-HQ~\cite{karras2017progressive} dataset. FT is a fine-tuned version of Lp3D, and ``frontal'' denotes frontalization of the source and driver.}
    \label{tab:abl}
    \vspace{-3mm}
\end{minipage}
\end{figure}

\begin{figure}[t]
    \centering
    \includegraphics[width=\linewidth]{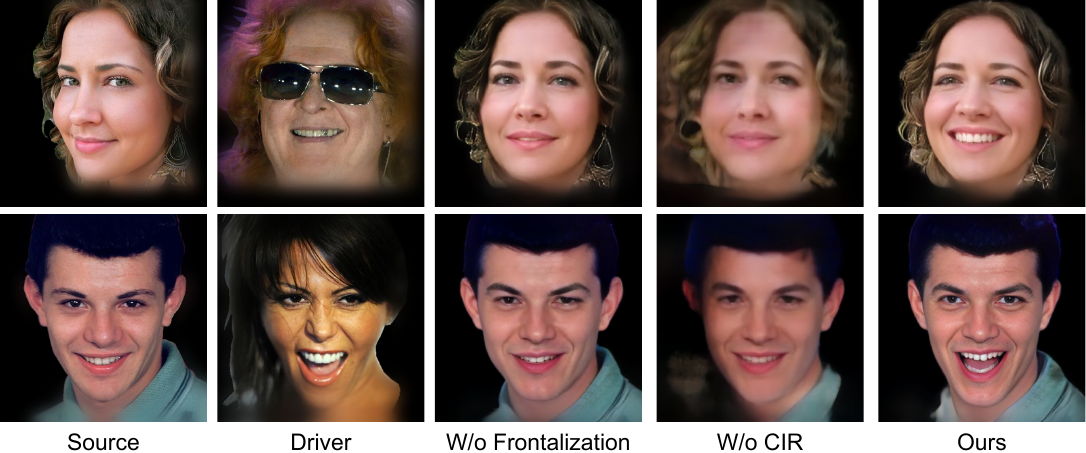}
    \vspace{-0.58cm}
    \caption{Ablation study for source and driver frontalization and cross identity regularization (CIR).}
    \label{fig:frontalizationcir}
\end{figure}

\begin{figure}
    \centering
    \includegraphics[width=0.95\linewidth]{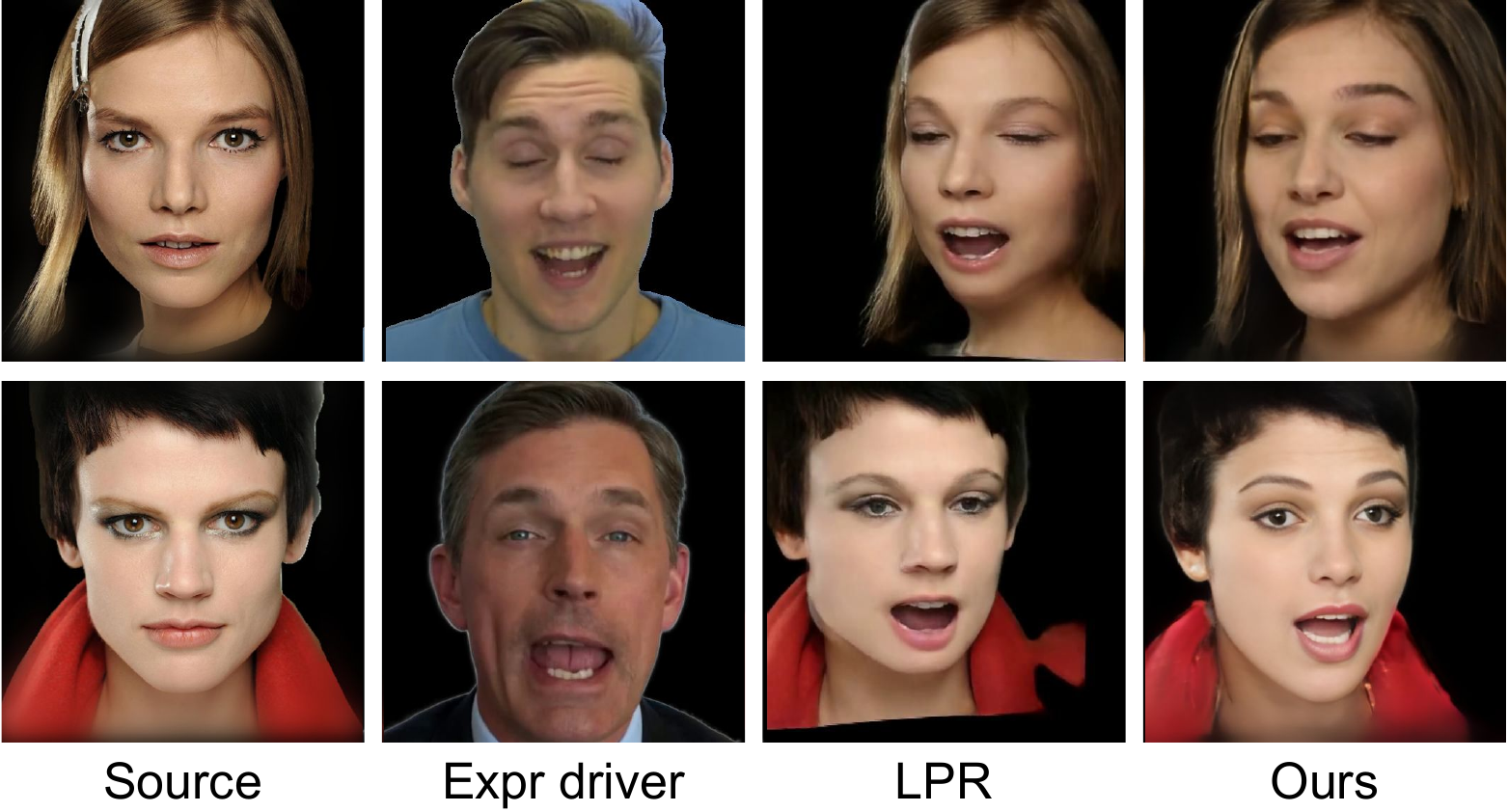}
    \vspace{-0.3cm}
    \caption{Qualitative comparison with LPR~\cite{li2023generalizable} method on the samples from HDTF~\cite{zhang2021flow} dataset.}
    \label{fig:lprcomp}
\end{figure}

\paragraph{Implementation Details.}

We train our model on CelebV-HQ dataset~\cite{zhu2022celebvhq} using 7 NVIDIA RTX A6000 ADA (50Gb memory each).
We use AdamW~\cite{Loshchilov2017DecoupledWD} to optimize the parameters with a learning rate of $10^{-4}$ and batch size of 28. 
The Lp3D finetuning takes 5 days for 500K iterations to converge. 
Training the expression module takes 2 days, and the iterative fine-tuning takes another 5 days. 
More training details, such as hyperparameter fine-tuning or architecture of the networks, can be found in the supplementary materials.

Unlike Lp3D, our method reenacts faces without re-lifting in 3D for every frame. For each driver, we perform only a single frontalization (0.0115 ms), one inference for expression encoding (0.0034 ms), and one tri-plane rendering at $128 \times 128$ resolution (0.0071 ms), and one neural upsampling (0.0099 ms). Each view runs at 31.9 fps on an Nvidia RTX 4090 GPU including I/O. More details on performance can be found in the supplemental materials.

We compare our method with state-of-the-art 3D-based~\cite{Khakhulin2022ROME,ma2023otavatar,hong2022headnerf} and 2D-based~\cite{drobyshev2022megaportraits,StyleHEAT_2022} models. 
For MegaPortraits~\cite{drobyshev2022megaportraits}, we use our own implementation that was trained on the CelebV-HQ dataset.
Similar to previous works, we evaluate our method using public benchmarks, including CelebA-HQ~\cite{karras2017progressive} and HDTF~\cite{zhang2021flow}. 
For CelebA-HQ, we split the data into two equal sets. Each set contains around 15K images. 
Then, we use one set as the source and the rest as driver images. 
For the HDTF dataset, we perform cross-reenactment by using the first frame of each video as source and 200 first frames of other videos as drivers, which is more than 60K data pairs.
Similarly, to evaluate self-reenactment, we also use the first frames of each video as sources and the rest of the same video as the driver. 
Furthermore, we also collected 100 face images on the internet and around 100 high-quality videos for qualitative comparison purposes.
We provide the video results in the supplementary materials.

\paragraph{Quantitative Comparisons.}
Given a source image $x_s$, a driver image $x_d$, and reenacted output $x_{s\rightarrow d}$ we use EMOCAv2 \cite{danvevcek2022emoca} to extract the FLAME \cite{li2017learning} expression coefficients of the prediction and the driver, as well as the shape coefficients of the source. We then compute 2 FLAME meshes using the predicted shape coefficients in world coordinates, one with the expression coefficients of the driver and one with the expression coefficients of the reenacted output. We measure the distance between the 2 meshes and denote this expression metric as ECMD.  
%\EZ{this part is completely missing me, which methods do you use to produce the meshes?}
%
Moreover, we also use cosine similarity between the embeddings of a face recognition network (CSIM)~\cite{zakharov2019few}, normalized average keypoint distance (NAKD)~\cite{bulat2017far}, perceptual image similarity (LPIPS) \cite{zhang2018unreasonable}, peak signal-to-noise ratio (PSNR), and structure similarity index measure (SSIM).

We provide quantitative comparisons on HDTF and CelebA-HQ datasets in \cref{tab:hdtf} and \cref{tab:celeba}, respectively, and show that our method outperforms existing methods on both datasets. 
We also note that our FID and CSIM scores are significantly more reliable than the others, while expression-based metrics such as NAKD and ECMD are either better or very close to the best baseline, w.r.t output quality, expression accuracy, and identity consistency.
 
\paragraph{Qualitative Results.}

\cref{fig:qual} and \cref{fig:wrinkles} showcase the qualitative results of cross-identity reenactment on in-the-wild images. Compared to the baselines ~\cite{hong2022headnerf,StyleHEAT_2022,drobyshev2022megaportraits,Khakhulin2022ROME}, our reenactment faithfully reconstructs intricate and complex elements, such as hairstyle, facial hair, glasses, and facial makeups. Furthermore, our method effectively generates realistic and fine-scale dynamic details that mach the driver's expressions including substantial head pose rotations. 
We also conduct a comparative analysis of our results with the current state-of-the-art 3D-aware method LPR~\cite{li2023generalizable} in \cref{fig:lprcomp}. Compared to LPR, our method achieves superior identity consistency.
We further refer to the supplemental video for a live demonstration of our holographic telepresence system and animated head reenactment results and comparisons, with and without disentangled poses.

\paragraph{Ablation Study.}

%\paragraph{Lp3D fine-tuning} 
We compare Lp3D with and without fine-tuning on the CelebA-HQ dataset in \cref{tab:abl} and show several examples in \cref{fig:lp3dfinetune}. 
Without fine-tuning on real data, our implementation of Lp3D fails to preserve the identity of the input image, resulting in a considerably lower CSIM score.
%
%\paragraph{Face frontalization} 
%
We also try without any facial frontalization in the expression module and instead use the source and driver images directly to calculate the expression tri-plane residual. 
We observe in~\cref{fig:frontalizationcir} that without face frontalization, the model completely ignores the expression of the driver and keeps the expression of the input source instead. 
We show in~\cref{tab:abl}, that facial frontalization leads to much better ECMD score.
%
%\paragraph{Cross identity regularization} 
We then measure the effectiveness of the GAN-based cross-identity regularization on the CelebA-HQ dataset, $\mathcal{L}_\text{CIR}$.
Without this loss, identity characteristics (hairstyle or color) can leak from the driver to the output. See column 4 in~\cref{fig:frontalizationcir}. 
\cref{tab:abl} also shows that cross-identity regularization can reduce identity leaking and improve the CSIM score.
%
%\paragraph{Iterative fine-tuning} 
Lastly, we have also attempted to train our model end-to-end using the same losses and optimization process instead of our proposed iterative fine-tuning. Even with a lower learning rate and the use of pre-trained Lp3D weights, we were unable to succeed.

%Lastly, instead of iterative fine-tuning, we train the model end to end using the same losses and optimization process. 
%
%However, we cannot get this model to work even with a lower learning rate and pre-trained Lp3D weights.

\paragraph{Limitations.}

\begin{figure}
    \centering
    \includegraphics[width=0.90\linewidth]{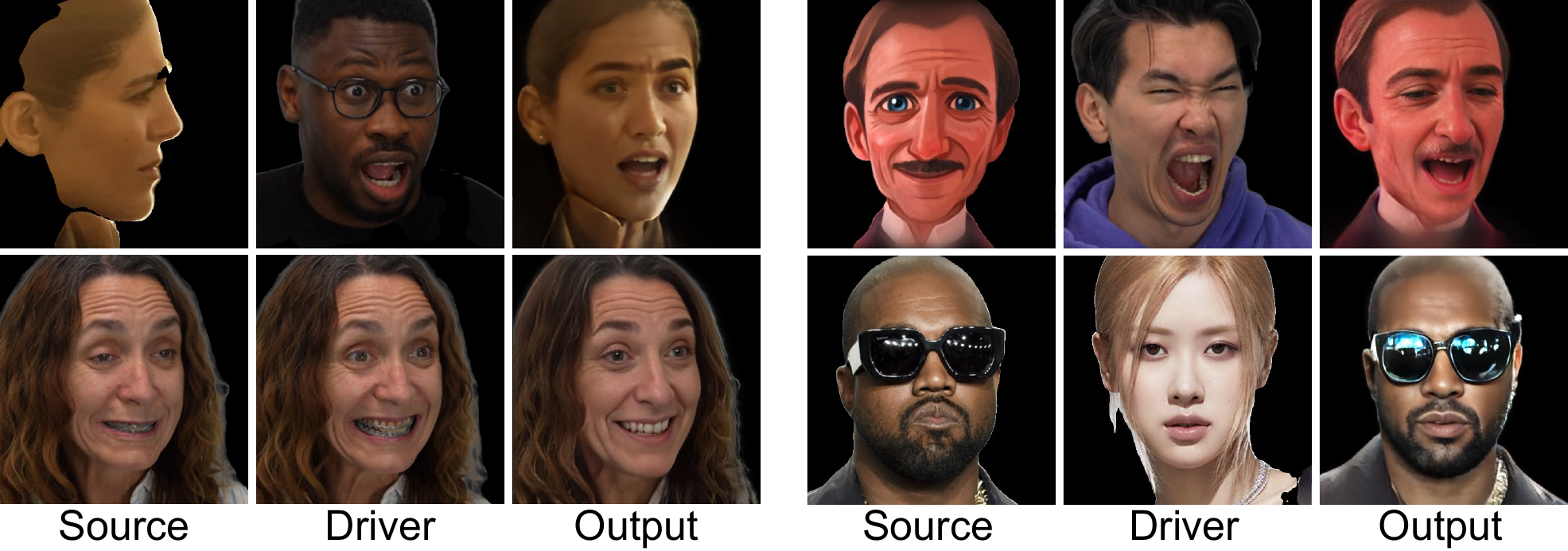}
    \vspace{-0.25cm}
    \caption{Failure cases of our method include side views in the source, extreme expressions, modeling of cartoonish characters and paintings, as well as modeling the reflections and semi-transparency of the eyewear.}
    \label{fig:limitation}
\end{figure}

Limitations of our approach are illustrated in Figure~\ref{fig:limitation}. For source images that are extremely side ways (i.e., over $90^\circ$), our method can produce a plausible frontal face, but the likeness cannot be guaranteed due to insufficient visibility. For very highly stylized portraits, such as cartoons, our framework often produces photorealistic facial elements such as teeth which can be inconsistent in style. Due to the dependence on training data volume and diversity, accessories such as dental braces or glasses may disappear or look different during synthesis. We believe that providing more and better training data can further improve the performance of our algorithm.

\section{Discussion}

We have demonstrated that a fully volumetric disentanglement of facial appearance and expressions is possible through a shared canonical tri-plane representation. 
In particular, an improved disentanglement also leads to higher fidelity and more robust head reenactment, when compared to existing methods that use linear face models for expressions, especially for non-frontal poses.
A critical insight of our approach is that head frontalization via 3D lifting is particularly effective for extracting features that can encode fine-scale details and expressions such as wrinkles and folds.
The resulting reenactment is also highly view-consistent for large angles, making our solution suitable for holographic displays.
We have also shown that the 3D lifting model can still be successfully trained with real data despite the fact that different frames with the same subject have varying facial expressions.
%
%We have also shown that fine-tuning a state-of-the-art 3D lifting method using in-the-wild data is possible via a loss function that combines both synthetic and real-world data terms. 
%
%While the synthetic data includes multi-view consistent portraits,  % when combined with the synthetic data term 
%
Without a fine-tuned 3D lifting model, our 3D-aware reenactment framework would struggle with preserving the identity of the source, especially for side views.
Our experiments indicate that our results achieve better visual quality and are more robust to extreme poses, which is validated via an extensive evaluation on multiple datasets.
%Our experiments also indicate that our results are not only visually superior but also more robust to highly challenging input examples where poses are non-frontal, and source faces have expressions, where we further support our claims using extensive quantitative evaluations on multiple datasets. 
\paragraph{Risks and Potential Misuse.} The proposed method is intended to promote avatar-based 3D communication. Nevertheless, our AI-based reenactment solution produces synthetic but highly realistic face videos from only a single photo, which could be hard to distinguish from a real person. Like deepfakes and other facial manipulation methods, potential misuse is possible and hence, we refer to the supplemental material for more discussions.
\paragraph{Future Work.}
%While we believe that further expressiveness (e.g., eye-gaze control) and robustness of our system can be achieved by incorporating more diverse data (faces, accessories, headwear, etc.) into the training, 
We are also interested in expanding our work to upper and full body reenactment, where hand gestures can be used for more engaging communication. 
To this end, we plan to investigate the use of canonical representations for human bodies, such as T-poses. 
%
% as our primary motivation to enable immersive 3D teleconferencing
As our primary motivation, we have showcased a solution using holographic displays for immersive 3D teleconferencing. However, we believe that our approach can also be extended to AR/VR HMD-based settings where full 360° head views are possible.
The recent work by An et al.~\cite{An_2023_CVPR} is a promising avenue for future exploration.
\clearpage
\setcounter{page}{1}
\maketitlesupplementary

\section{Training Details}
\begin{figure}[ht]
    \centering
    \includegraphics[width=0.9\linewidth]{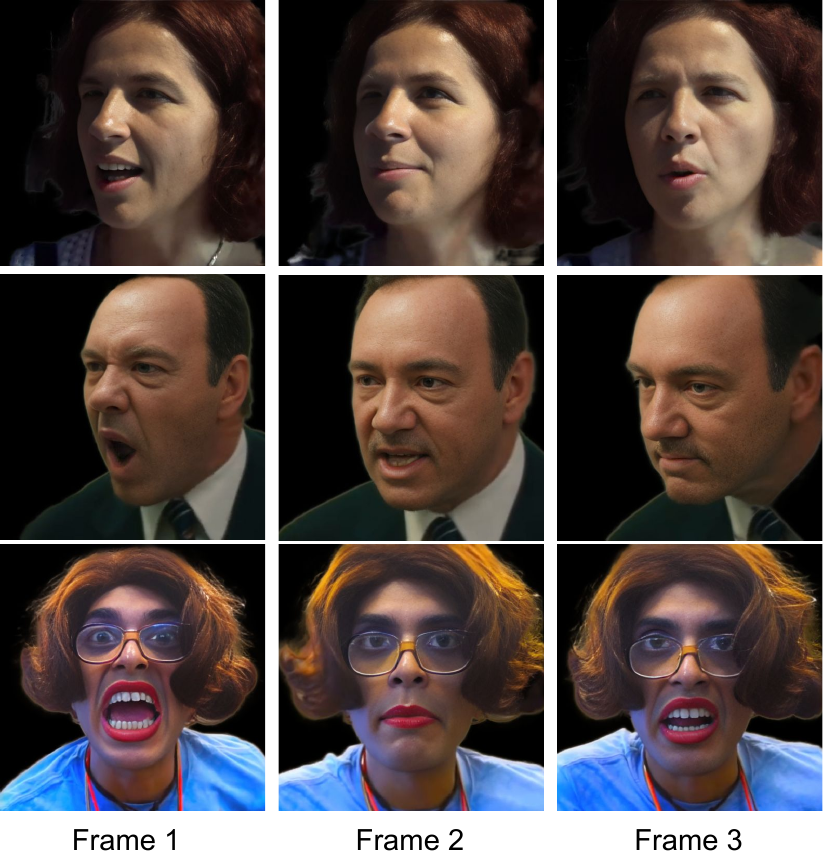}
    \caption{Some examples of our training data extracted from the CelebV-HQ dataset \cite{zhu2022celebvhq}}
    \label{fig:celebvhq}
\end{figure}

\paragraph{Training Data.}
We fine-tune Lp3D using CelebV-HQ dataset~\cite{zhu2022celebvhq}. For the expression modules, we also use the CelebV-HQ dataset but adopt an expression re-sampling process to make the expressions of the sources and drivers during training more different. Specifically, for a given video, we use EMOCA~\cite{danvevcek2022emoca} to reconstruct the mesh of every frame without the head pose. Let these obtained meshes be $\{M_1, M_2, ..., M_n\}$, we first pick two frames $x^*$ and $y^*$ such that the distance between their meshes are maximized:
\begin{align*}
    x^*, y^* = \argmax_{x, y} \|M_x - M_y\|_2.
\end{align*}
Then we pick the third frame $z^*$ such that:
\begin{align*}
    z^* = \argmax_z min\left(\|M_{x^*} - M_{z}\|, \|M_{y^*} - M_{z}\|\right).
\end{align*}
We use this frame selection process for all the videos in the CelebV-HQ dataset~\cite{zhu2022celebvhq} and use the re-sampled frames to train the expression modules. A few examples from this selection process are shown in \cref{fig:celebvhq}.

\paragraph{Driver Augmentation.} To prevent identity leaking from the driver to the output, we apply several augmentations to the frontalized driver images, including: (1) Kornia color jiggle\footnote{\url{https://kornia.readthedocs.io/en/latest/augmentation.module.html\#kornia.augmentation.ColorJiggle}} with parameters for brightness, contrast, saturation, hue set to 0.3, 0.4, 0.3, and 0.4, respectively; (2) random channel shuffle; (3) random warping\footnote{\url{https://github.com/deepfakes/faceswap/blob/a62a85c0215c1d791dd5ca705ba5a3fef08f0ffd/lib/training/augmentation.py\#L318}}; and (4) random border masking with the mask ratio uniformly sampled from 0.1 to 0.3. During testing, we removed all the augmentations except the random masking and fixed the mask ratio to 0.25. This random masking greatly improves the consistency in the output, especially for border regions. In addition, since we mask the border with a fixed rate, we can modify the renderer to only generate the center of the frontalized driver and further improve the performance.

\paragraph{Architecture Details.} Our architecture design is inspired by Lp3D~\cite{trevithick2023real}. Specifically, for $\mathbf{E}_s$ and $\mathbf{E}_d$, we use two separate DeepLabV3~\cite{chen2017rethinking} with all normalization layers removed. Since the triplane already captures deep 3D features of the source, we adopt a simple convolutional network for $\mathbf{E}_t$, which is given in \cref{tab:etarch}. Recall that:
\begin{align*}
    F = F_s \oplus F_d \oplus F_t
\end{align*}
For the final transformer that is applied on the concatenations of the feature maps $F$, we use a slight modification of $\mathbf{E}_{\text{low}}$ (light-weight version) in Lp3D~\cite{trevithick2023real}. The architecture of this module is given in \cref{tab:vitarch} where block used is the transformer block in SegFormer~\cite{xie2021segformer}. As mentioned in our paper, we use a pretrained GFPGAN as the super-resolution module. This module is loaded from a public pretrained weight GFPGAN~v1.4~\cite{wang2021towards} and fine-tuned end-to-end with the network.

\begin{table}
    \centering
    \begin{tabular}{|l|}
    \hline
        Conv2d(96, 96, kernel\_size=3, stride=2, padding=1)\\
        ReLU()\\
        Conv2d(96, 96, kernel\_size=3, stride=1, padding=1)\\
        ReLU()\\
        Conv2d(96, 128, kernel\_size=3, stride=2, padding=1)\\
        ReLU()\\
        Conv2d(128, 128, kernel\_size=3, stride=1, padding=1)\\
        ReLU()\\
        Conv2d(128, 128, kernel\_size=3, stride=1, padding=1)\\
    \hline
    \end{tabular}
    \caption{Architecture of $E_T$}
    \label{tab:etarch}
\end{table}

\begin{table}
    \centering
    \begin{tabular}{|l|}
    \hline
        PatchEmbed(64, patch=3, stride=2, in=640, embed=1024)\\
        Block(dim=1024, num\_heads=4, mlp\_ratio=2, sr\_ratio=1)\\
        Block(dim=1024, num\_heads=4, mlp\_ratio=2, sr\_ratio=1)\\
        PixelShuffle(upscale\_factor=2)\\
        upsample(scale\_factor=2, mode=bilinear)\\
        Conv2d(256, 128, kernel\_size=3, stride=1, padding=1)\\
        ReLU()\\
        upsample(scale\_factor=2, mode=bilinear)\\
        Conv2d(128, 128, kernel\_size=3, stride=1, padding=1)\\
        ReLU()\\
        Conv2d(128, 96, kernel\_size=3, stride=1, padding=1)\\
    \hline
    \end{tabular}
    \caption{Architecture of the transformer network used in the expression module.}
    \label{tab:vitarch}
\end{table}

\paragraph{Training Losses.}
To train the model used in our experiments, we set $\lambda_{\text{syn}} = 0.1, \lambda_{\text{tri}} = 0.01$, and $\lambda_{\text{CIR}} = 0.01$. For GAN-based losses, we use hinge loss~\cite{lim2017geometric} with projected discriminator~\cite{sauer2021projected}.

\section{Implementation Details for Holographic Display System}
We implement our model on a Looking Glass monitor $32^{"}$\footnote{\url{https://lookingglassfactory.com/looking-glass-32}}. To visualize results on a holographic display, we must render multiple views for each frame using camera poses with a yaw angle that spans the range from $-17.5^\circ$ to $17.5^\circ$. In our case, we find that using 24 views is sufficient for the user experience. While our model can run at 32FPS using a single NVIDIA RTX 4090 on a regular monitor, which only requires a single view at a time, it cannot run in real-time when rendering 24 views simultaneously. %With a single NVIDIA RTX 6000 ADA GPU, the triplane renderer can render a single view image at 140~FPS. 
Thus, to achieve real-time performance for the Looking Glass display, we ran the holographic telepresence demo on seven NVIDIA RTX 6000 ADA GPUs. 

We parallelize the rendering process to four GPUs, so each one needs to render six views in a batch. We dedicate one GPU for driving image pre-processing and another one for disentangled tri-plane estimation. We use the last GPU to run the looking-glass display itself. This setup results in 25~FPS for the whole application. We showcase the results rendered on the holographic display in the supplementary videos.

\section{Additional Comparisons with LPR \cite{li2023generalizable}}
\begin{figure}
    \centering
    \includegraphics[width=0.95\linewidth]{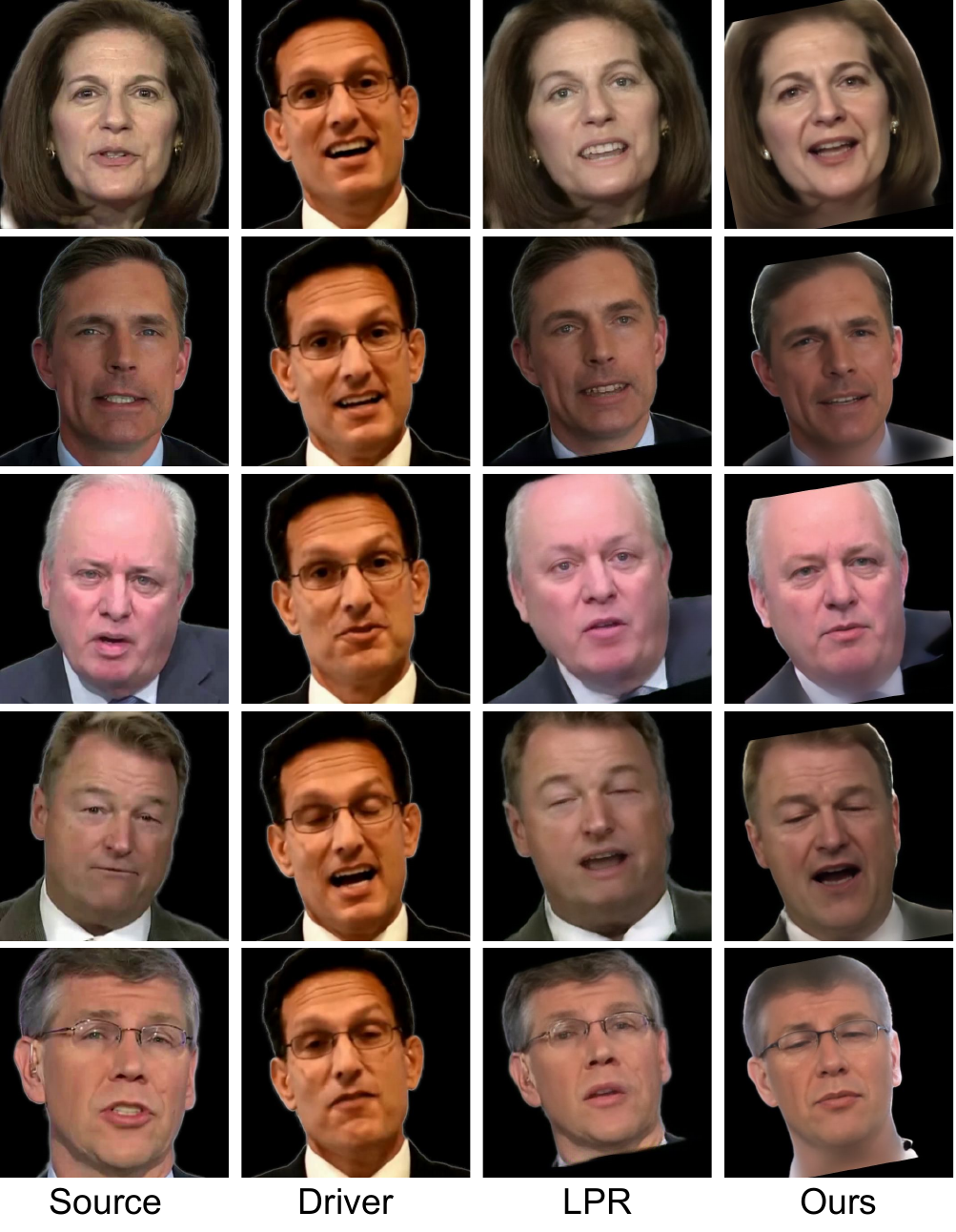}
    \caption{Qualitative comparisons with LPR \cite{li2023generalizable} on HDTF dataset.}
    \label{fig:lprhdtfqual}
\end{figure}

In this section, we compare our method with the current state-of-the-art in 3D aware one-shot head reenactment, LPR \cite{li2023generalizable} using their test data from HDTF \cite{zhang2021flow} and CelebA-HQ datasets \cite{karras2017progressive}. In particular, for CelebA-HQ, they use even-index frames as sources and odd-index frames as drivers, while in contrast, in our experiment section, we use the first half as sources and the rest as drivers. For the HDTF dataset, they use a single driver (WRA\_EricCantor\_000) and the first frame of each video as source image. Compared to our split, this reduces the diversity in the driver images. We provide the comparison results in \cref{tab:lprhdtfquan} and \cref{tab:lprcelebaquan}. The ECMD scores on both datasets show that our method is more accurate in transferring expression from the driver to the source images. On the HDTF dataset, our results have much higher CSIM. Our FID score is better than LPR \cite{li2023generalizable} on CelebA-HQ but worse on the HDTF dataset. We found that the HDTF's ground-truth images have poor quality while our outputs are higher in quality; this mismatch causes our FID to be unimpressive on this dataset. Hence, this FID arguably does not correctly reflect the performance of our model. According to the qualitative examples in \cref{fig:lprhdtfqual}, our method captures the driver's expression more accurately than LPR. However, we note that our quality is even higher than the input, as can be observed in \cref{fig:lprhdtfqual}.

\begin{figure}[ht]
    \centering
    \includegraphics[width=0.95\linewidth]{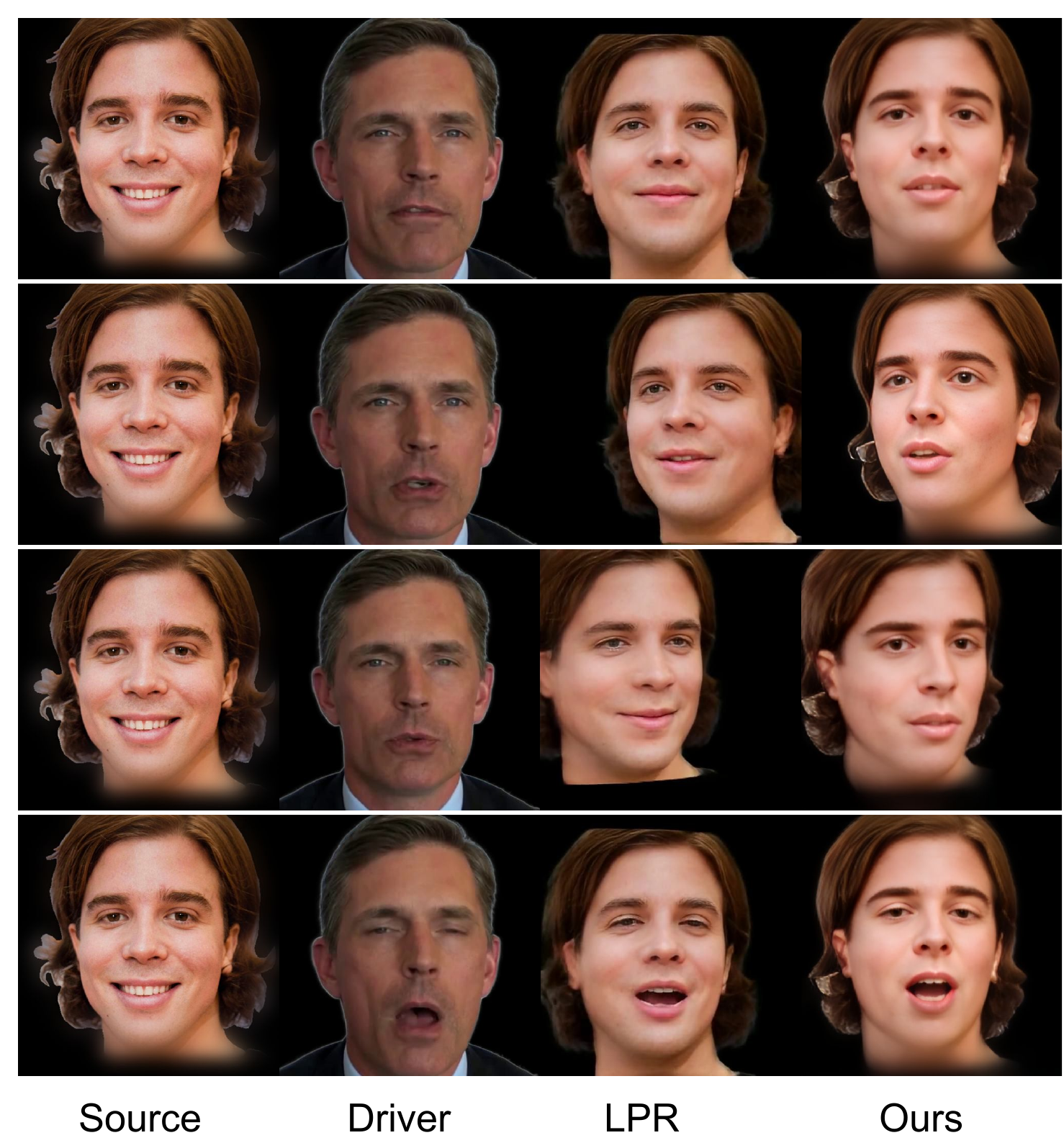}
    \caption{Novel view synthesis comparison with LPR. In this example, LPR fails to remove the smiling expression from the source while our method successfully transfer the expression from the driver to the source due to better disentanglement.}
    \label{fig:lprsmile}
\end{figure}

\begin{figure*}[ht]
    \centering
    \includegraphics[width=0.95\linewidth]{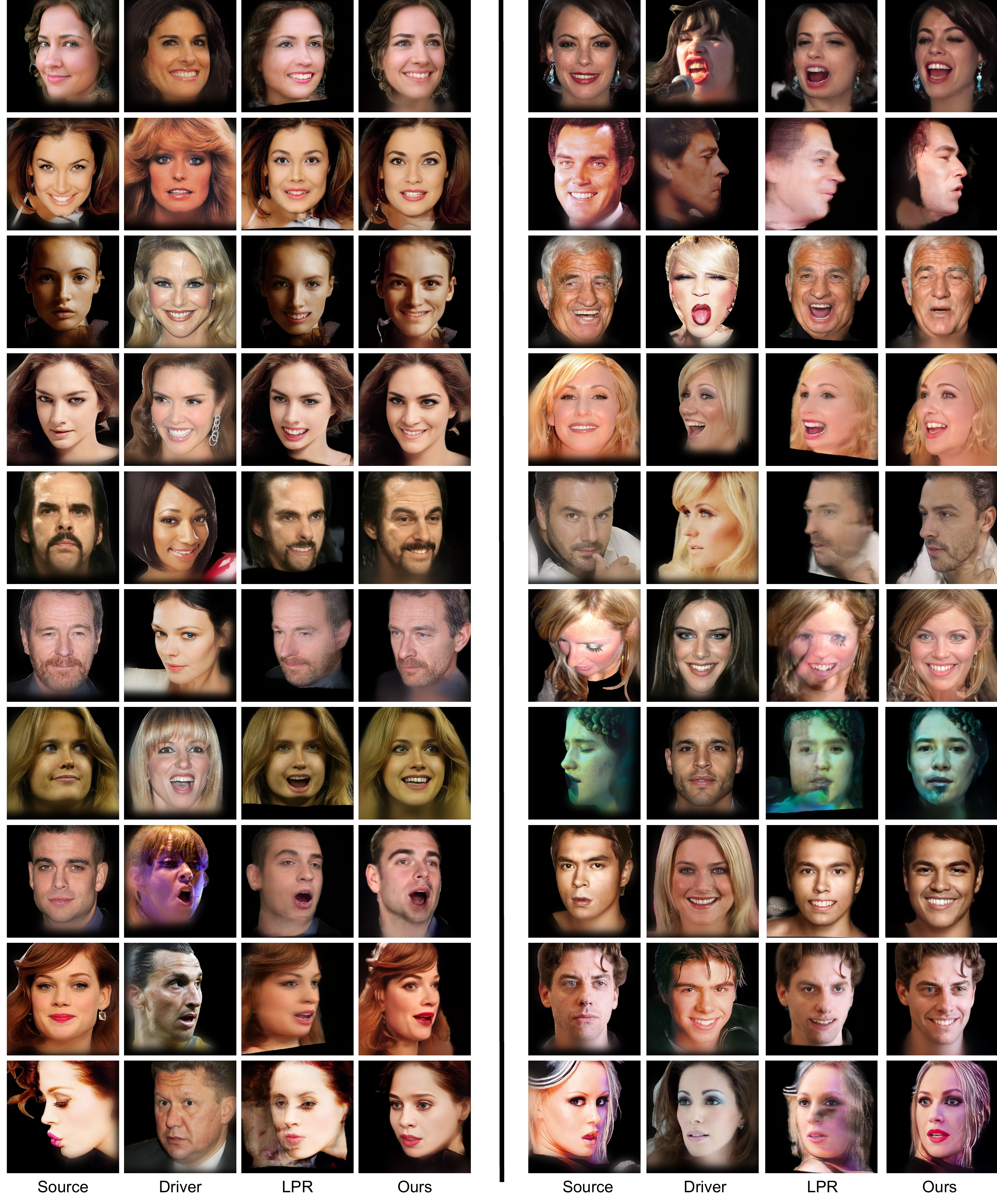}
    \caption{Qualitative comparisons with LPR \cite{li2023generalizable} on CelebA-HQ dataset.}
    \label{fig:lprcelebaqual}
\end{figure*}

We also provide extensive qualitative comparisons in \cref{fig:lprcelebaqual} and \cref{fig:lprhdtfqual}. The expression of our output images is more realistic and faithful to the driver, which is particularly more visible in the mouth/teeth/jaw region, as well as for driver or source side views. Notably, in \cref{fig:lprsmile}, it can be observed that LPR fails to remove the smiling from the source, resulted in inaccurate expression in the reenacted output while our method can still successfully transfer the expression from the driver to the source image.

% \section{Typo Corrections.}
% There is a typo in the concatenation equation in sec. 3.2. The correct one is:
% \begin{align*}
%     F = F_s \oplus F_d \oplus F_T
% \end{align*}

\begin{table}[ht]
    \centering
    \begin{tabular}{l|ccc}
        \multirow{2}{*}{Method} & \multicolumn{3}{c}{Cross-reenactment}\\
        \cline{2-4}
         & CSIM  & ECMD & FID \\
         \hline
         LPR \cite{li2023generalizable} & 0.531 & 0.912 & \textbf{25.26}\\
         Ours & \textbf{0.774} & \textbf{0.860} & 54.15\\
    \end{tabular}
    \caption{Quantitative comparisons with LPR \cite{li2023generalizable} on HDTF dataset using the test split proposed in \cite{li2023generalizable}.}
    \label{tab:lprhdtfquan}
\end{table}

\begin{table}[ht]
    \centering
    \begin{tabular}{l|ccc}
        \multirow{2}{*}{Method} & \multicolumn{3}{c}{Cross-reenactment}\\
        \cline{2-4}
         & CSIM  & ECMD & FID \\
         \hline
         LPR \cite{li2023generalizable} & \textbf{0.643} & 0.483 & 47.39\\
         Ours & 0.628 & \textbf{0.473} & \textbf{34.27}\\
    \end{tabular}
    \caption{Quantitative comparisons with LPR \cite{li2023generalizable} on CelebA-HQ dataset  using the test split proposed in \cite{li2023generalizable}.}
    \label{tab:lprcelebaquan}
\end{table}

\section{Additional Qualitative Comparisons}
\begin{figure*}
    \centering
    \includegraphics[width=0.85\linewidth]{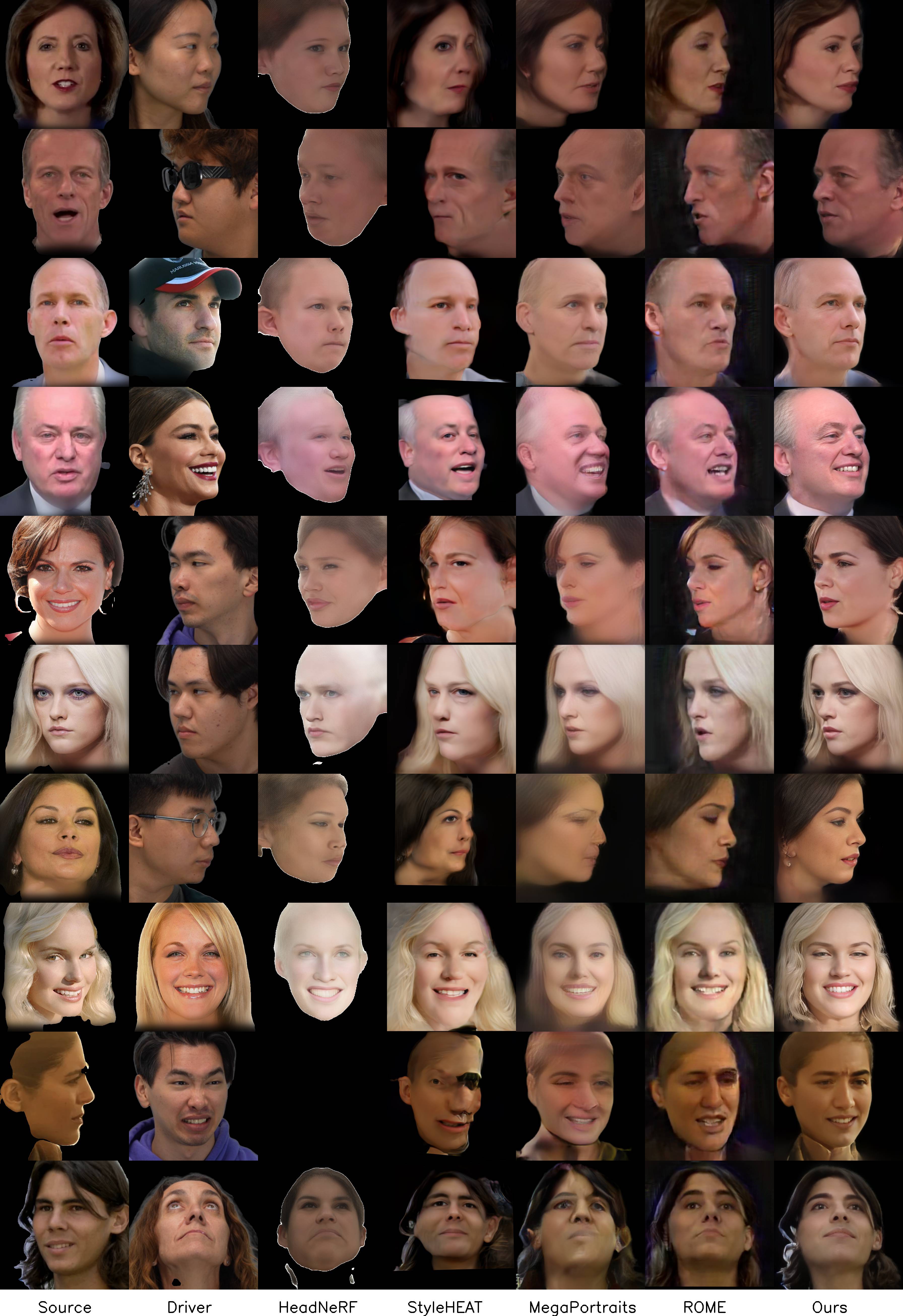}
    \caption{Qualitative results on various datasets.}
    \label{fig:supqal01}
\end{figure*}

\begin{figure*}
    \centering
    \includegraphics[width=0.85\linewidth]{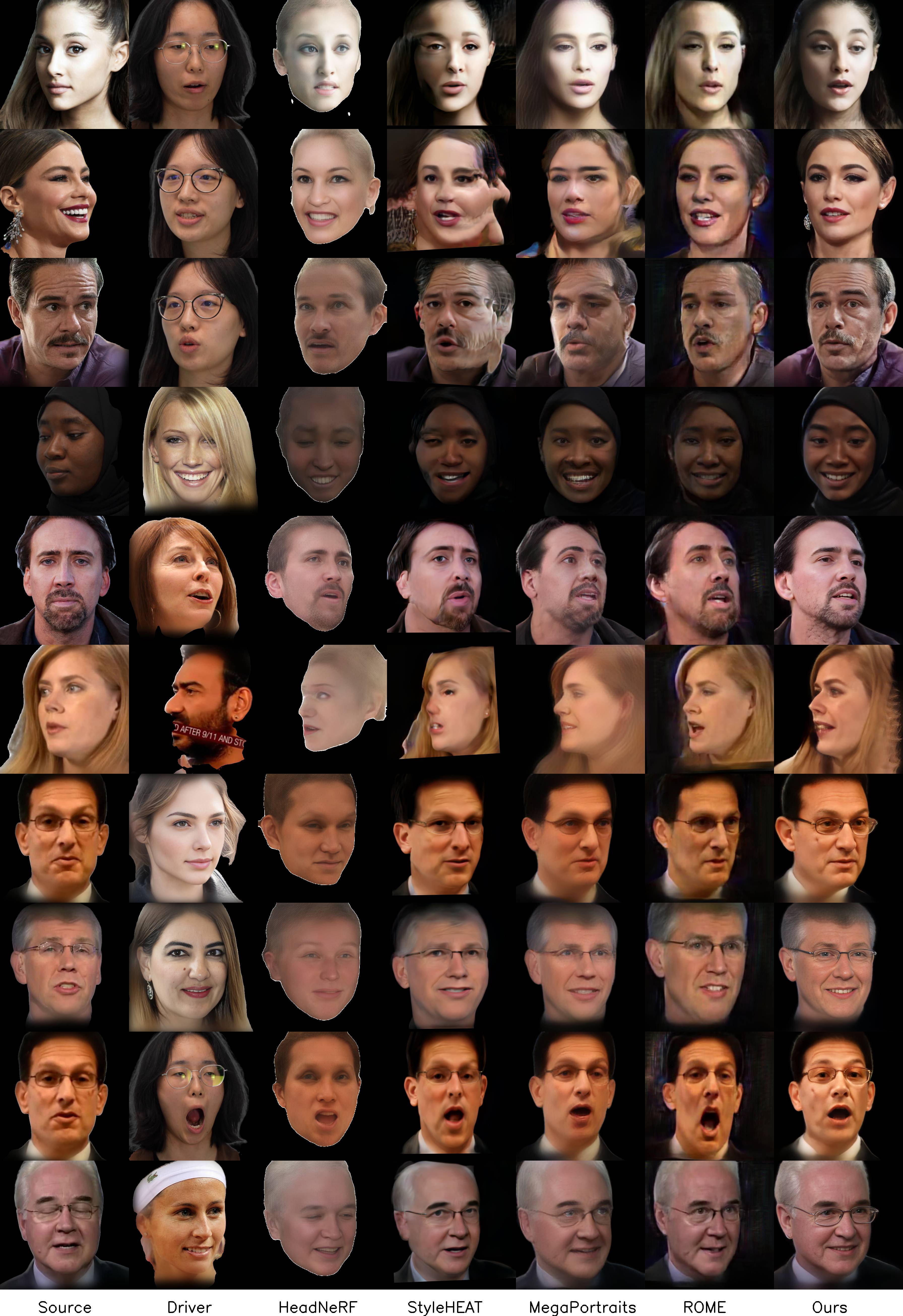}
    \caption{Qualitative results on various datasets.}
    \label{fig:supqal02}
\end{figure*}

\begin{figure*}
    \centering
    \includegraphics[width=0.85\linewidth]{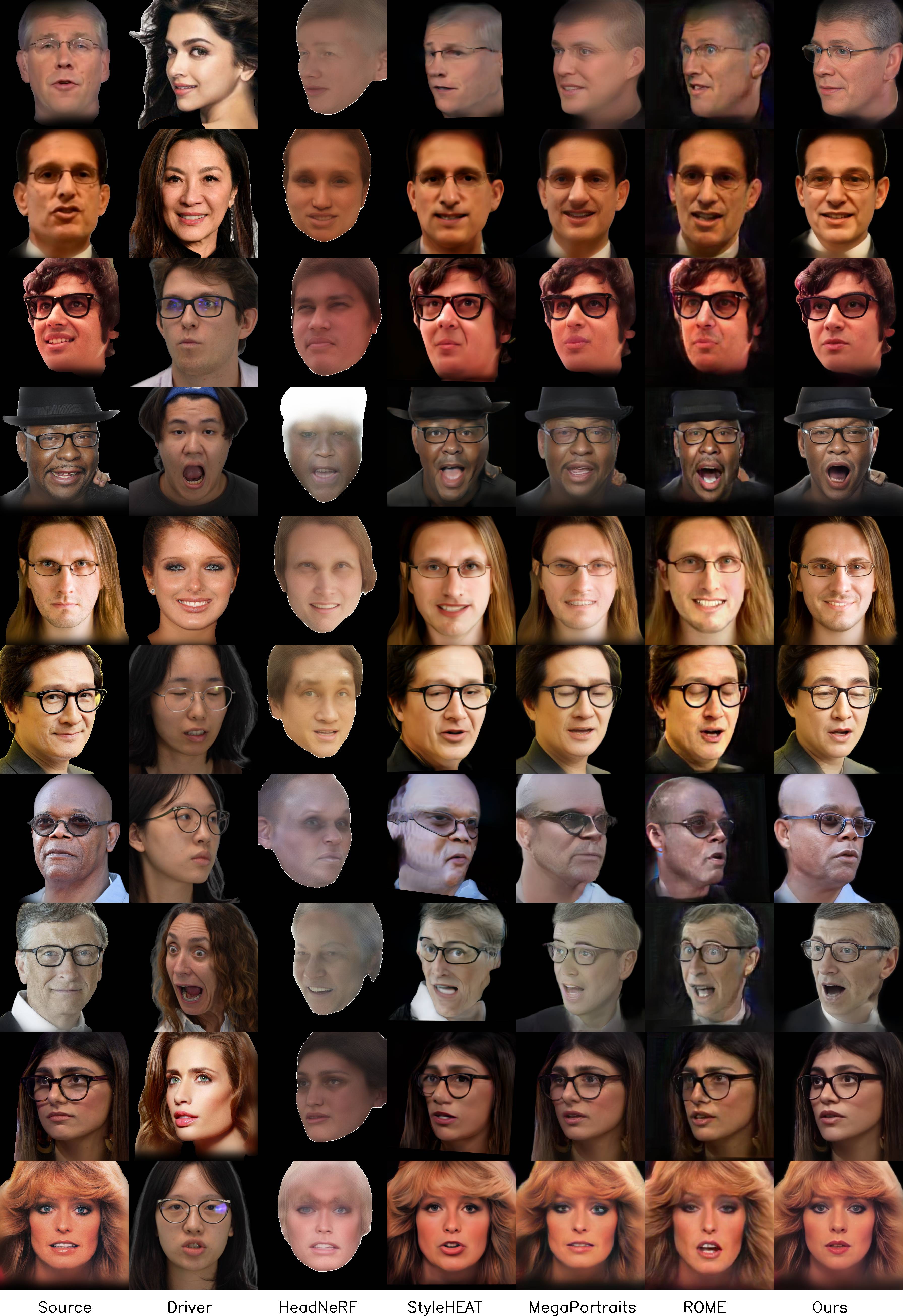}
    \caption{Qualitative results on various datasets.}
    \label{fig:supqal03}
\end{figure*}

\begin{figure*}
    \centering
    \includegraphics[width=0.85\linewidth]{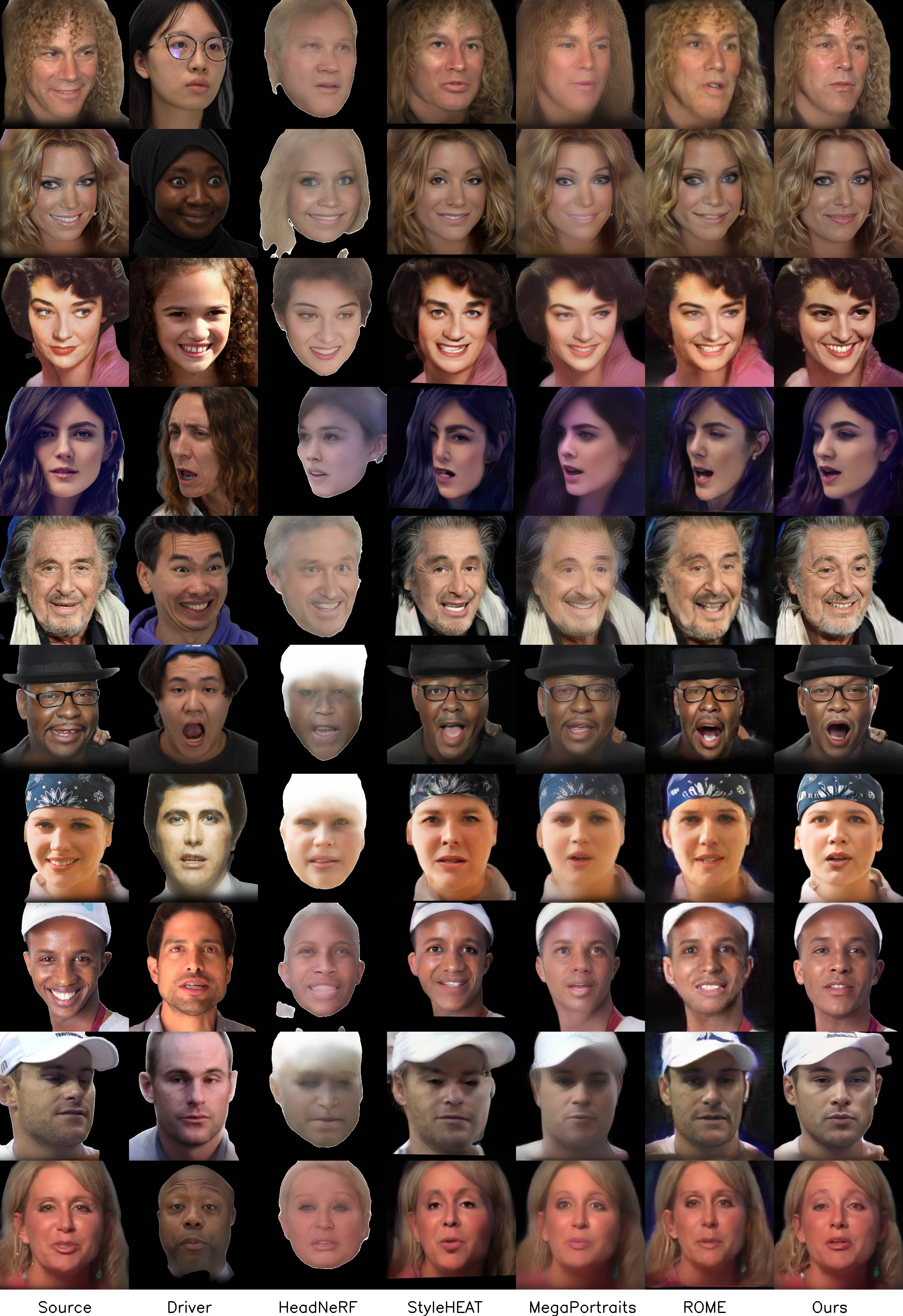}
    \caption{Qualitative results on various datasets.}
    \label{fig:supqal04}
\end{figure*}

\begin{figure*}
    \centering
    \includegraphics[width=0.85\linewidth]{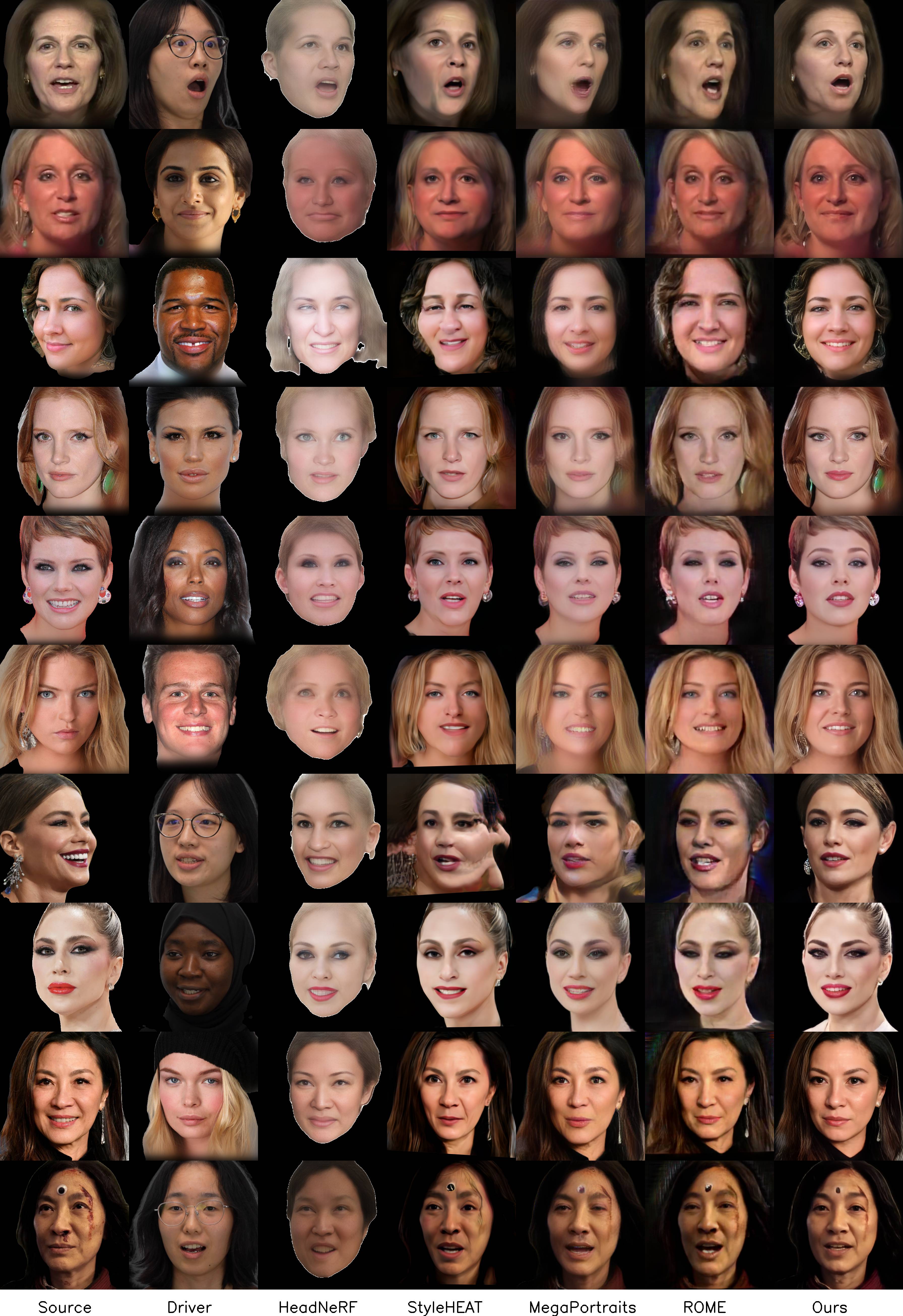}
    \caption{Qualitative results on various datasets.}
    \label{fig:supqal05}
\end{figure*}

\begin{figure*}
    \centering
    \includegraphics[width=0.85\linewidth]{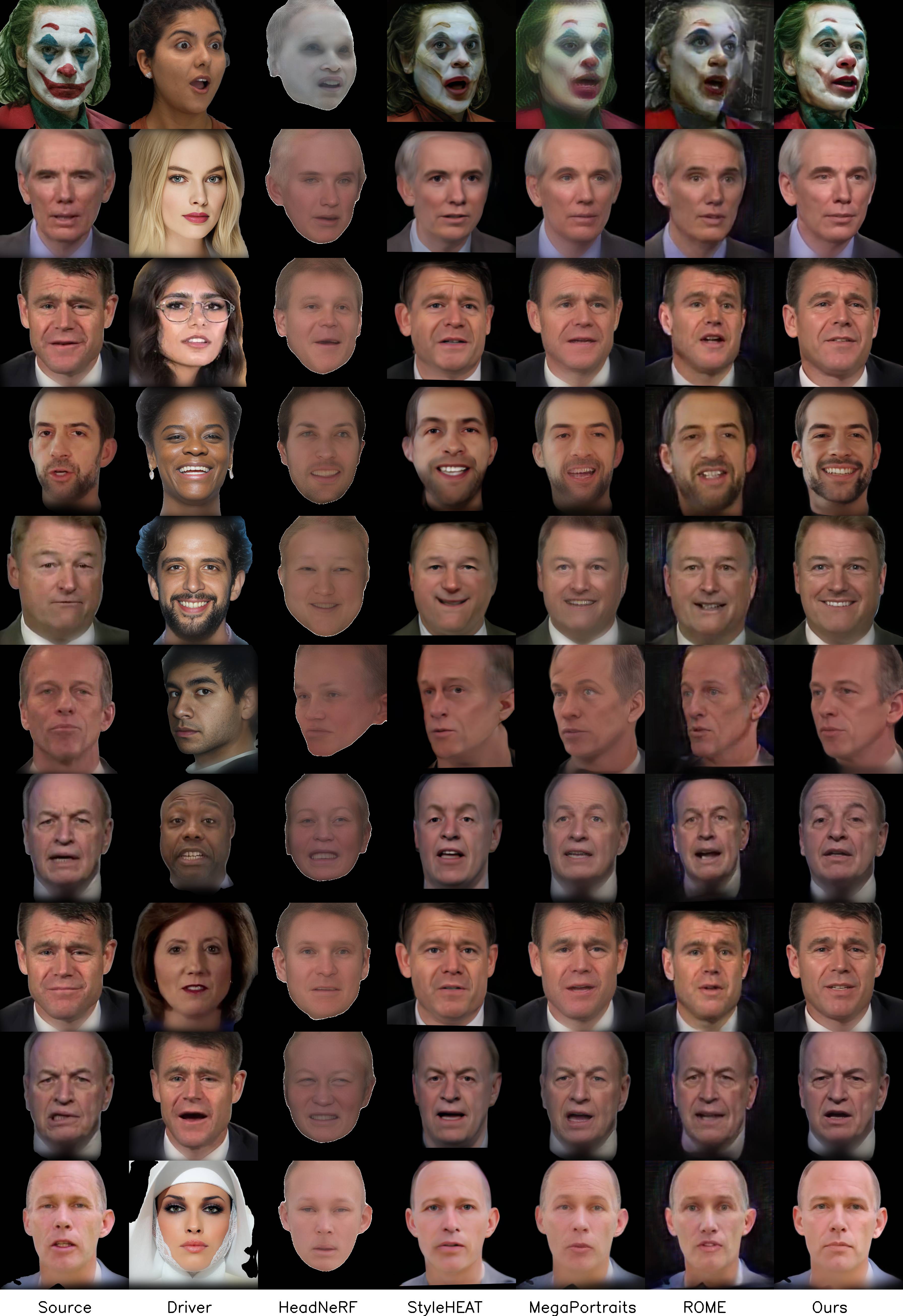}
    \caption{Qualitative results on various datasets.}
    \label{fig:supqal06}
\end{figure*}

\begin{figure*}
    \centering
    \includegraphics[width=0.85\linewidth]{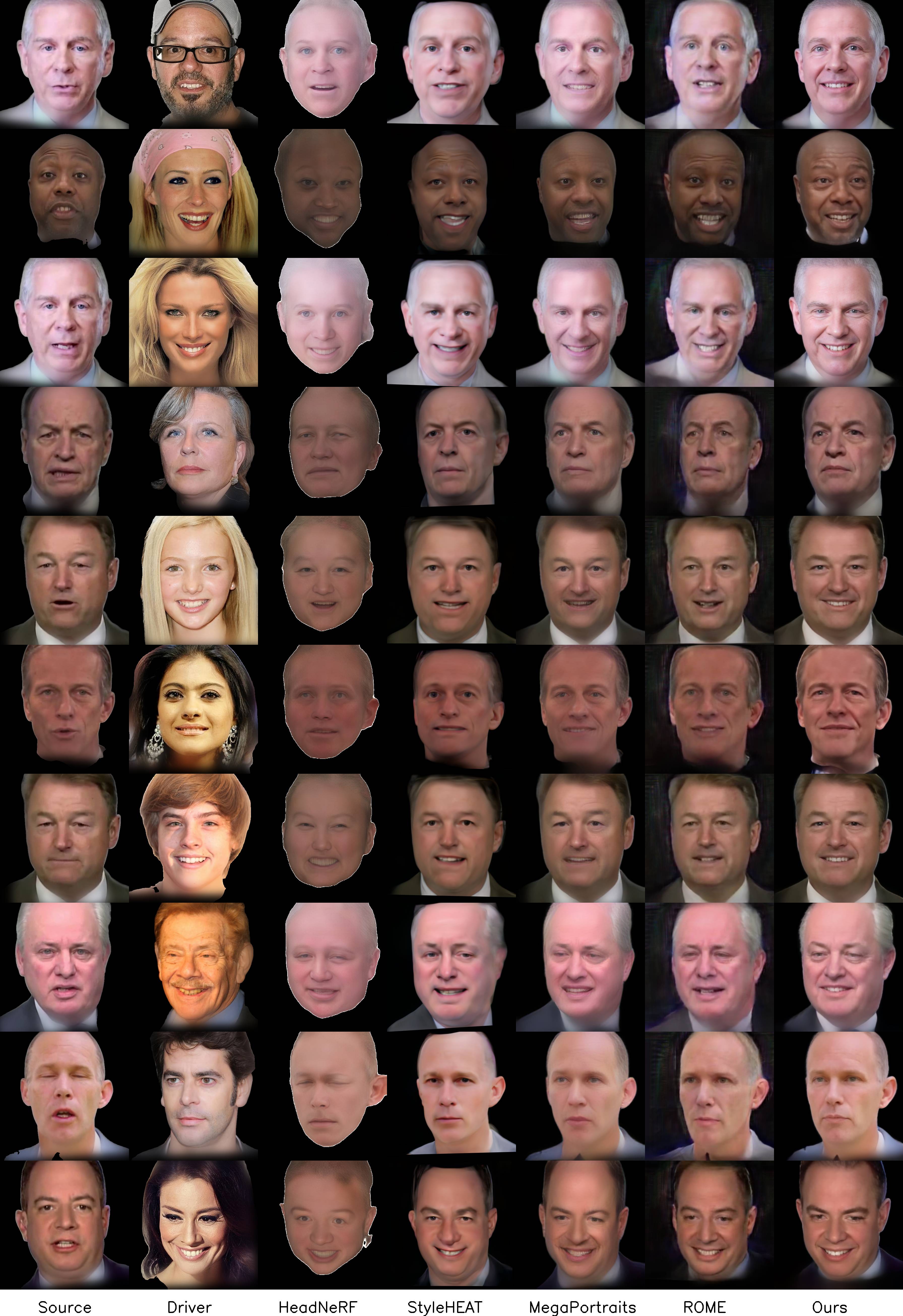}
    \caption{Qualitative results on various datasets.}
    \label{fig:supqal07}
\end{figure*}

\begin{figure*}
    \centering
    \includegraphics[width=0.85\linewidth]{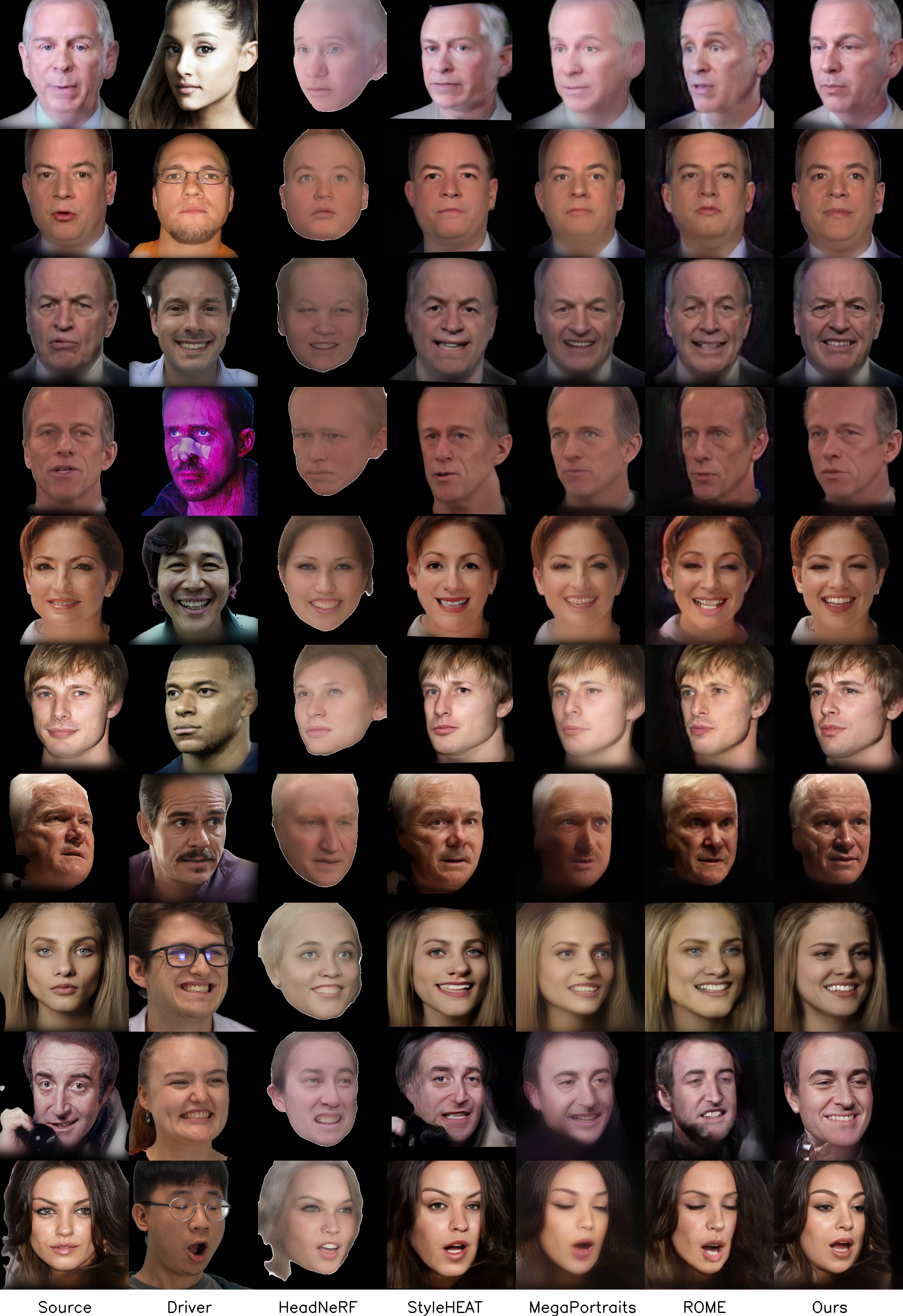}
    \caption{Qualitative results on various datasets.}
    \label{fig:supqal08}
\end{figure*}

\begin{figure*}
    \centering
    \includegraphics[width=0.85\linewidth]{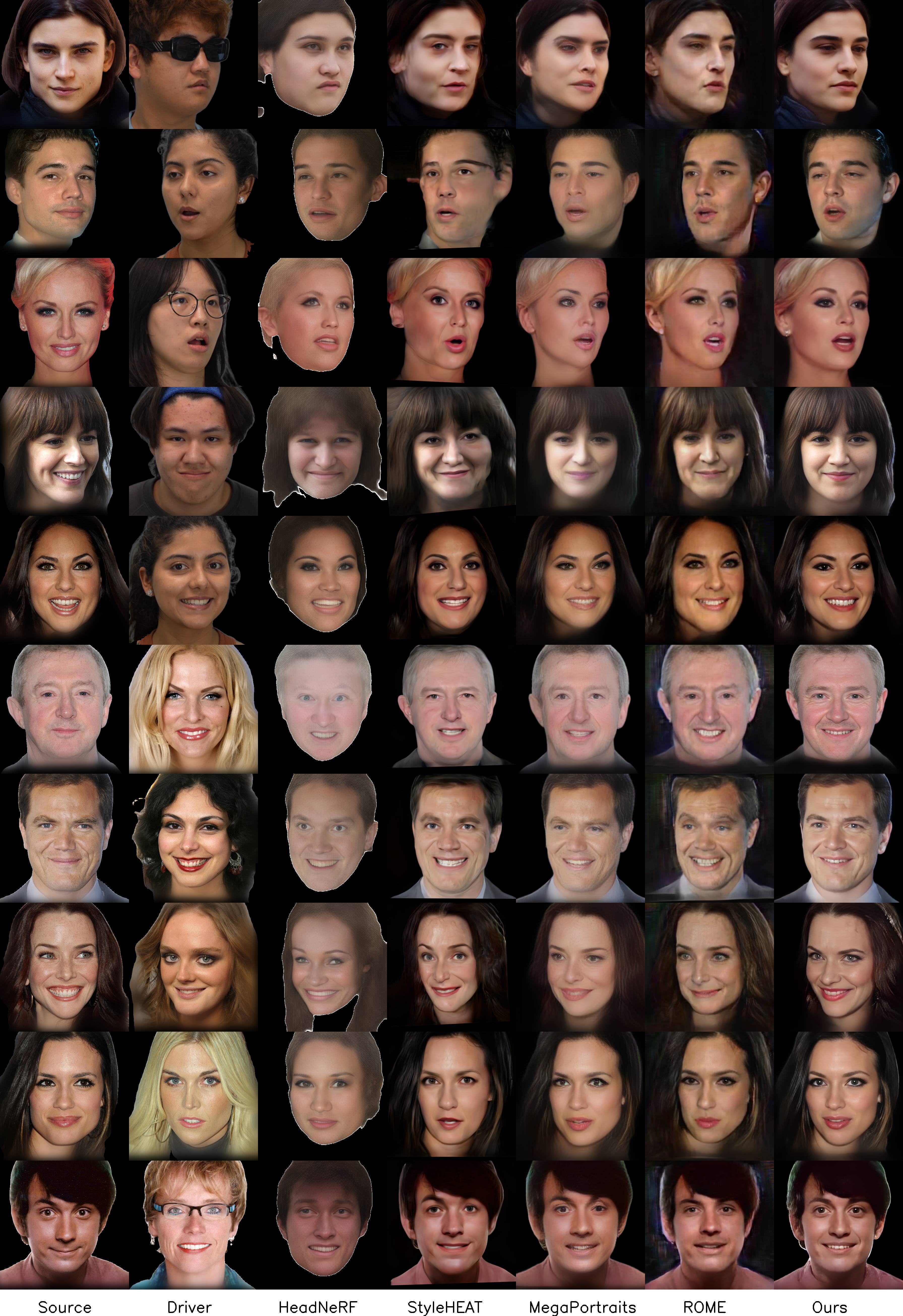}
    \caption{Qualitative results on various datasets.}
    \label{fig:supqal09}
\end{figure*}

\begin{figure*}
    \centering
    \includegraphics[width=0.85\linewidth]{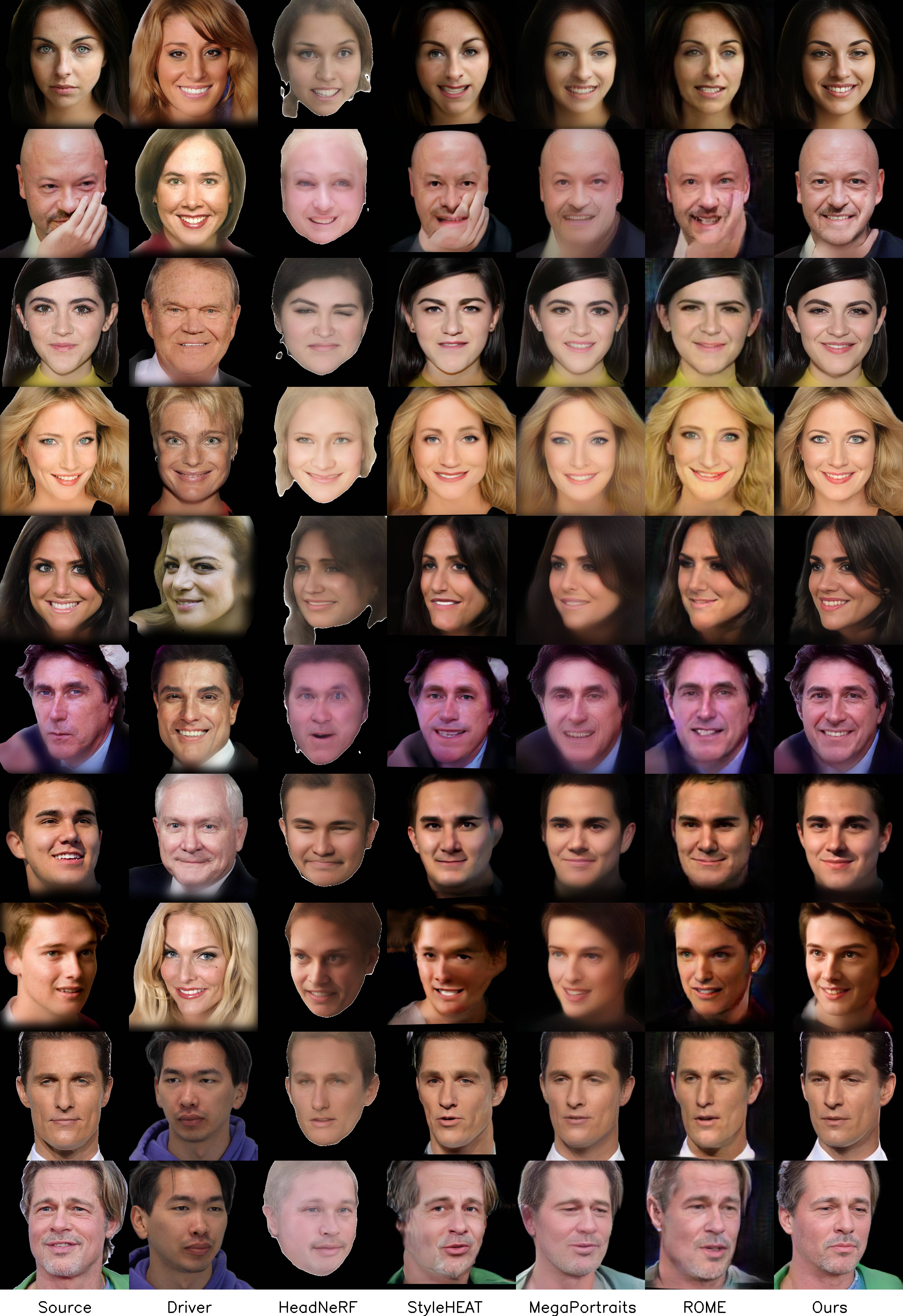}
    \caption{Qualitative results on various datasets.}
    \label{fig:supqal10}
\end{figure*}

\begin{figure*}
    \centering
    \includegraphics[width=0.85\linewidth]{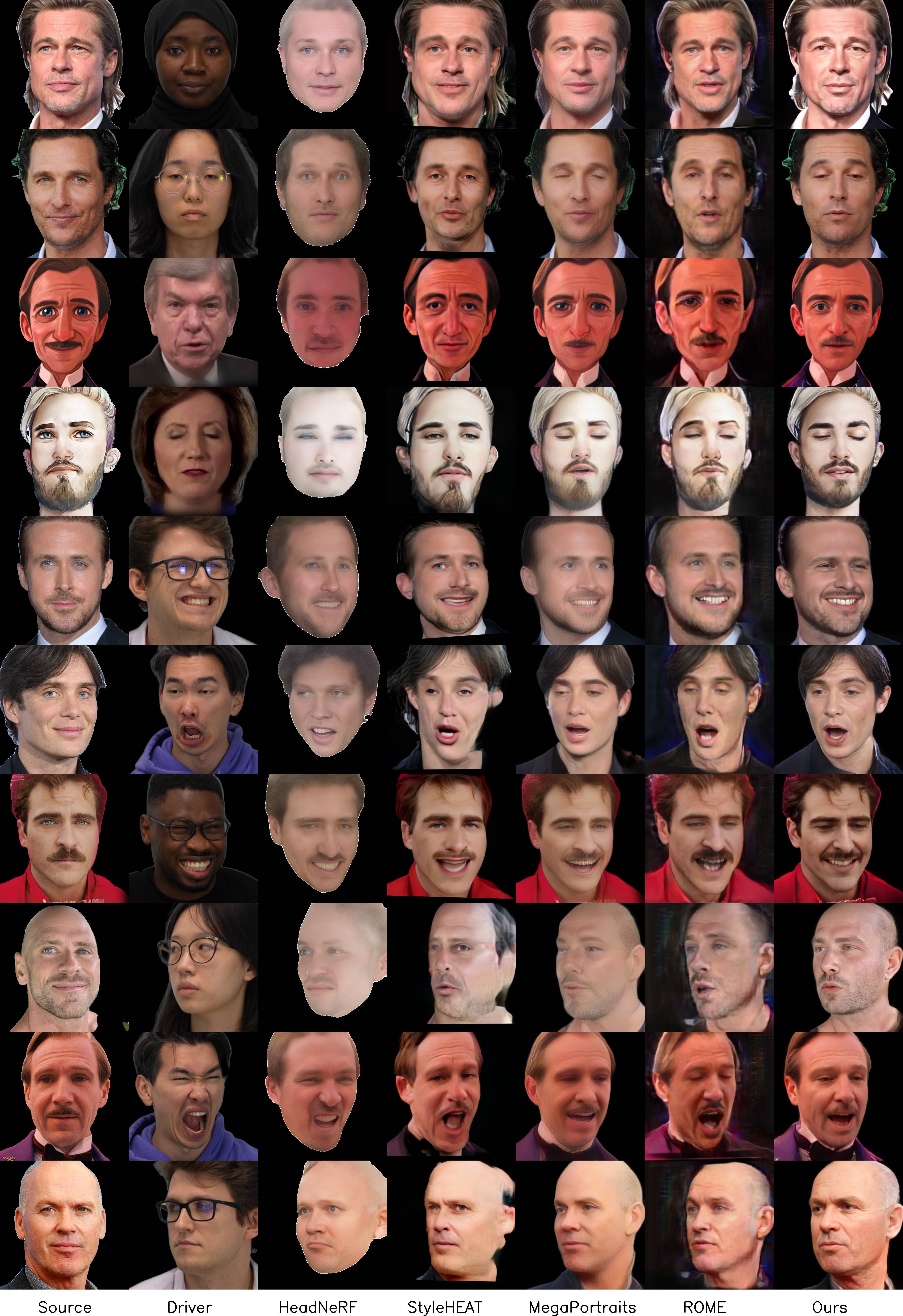}
    \caption{Qualitative results on various datasets.}
    \label{fig:supqal11}
\end{figure*}

\begin{figure*}
    \centering
    \includegraphics[width=0.85\linewidth]{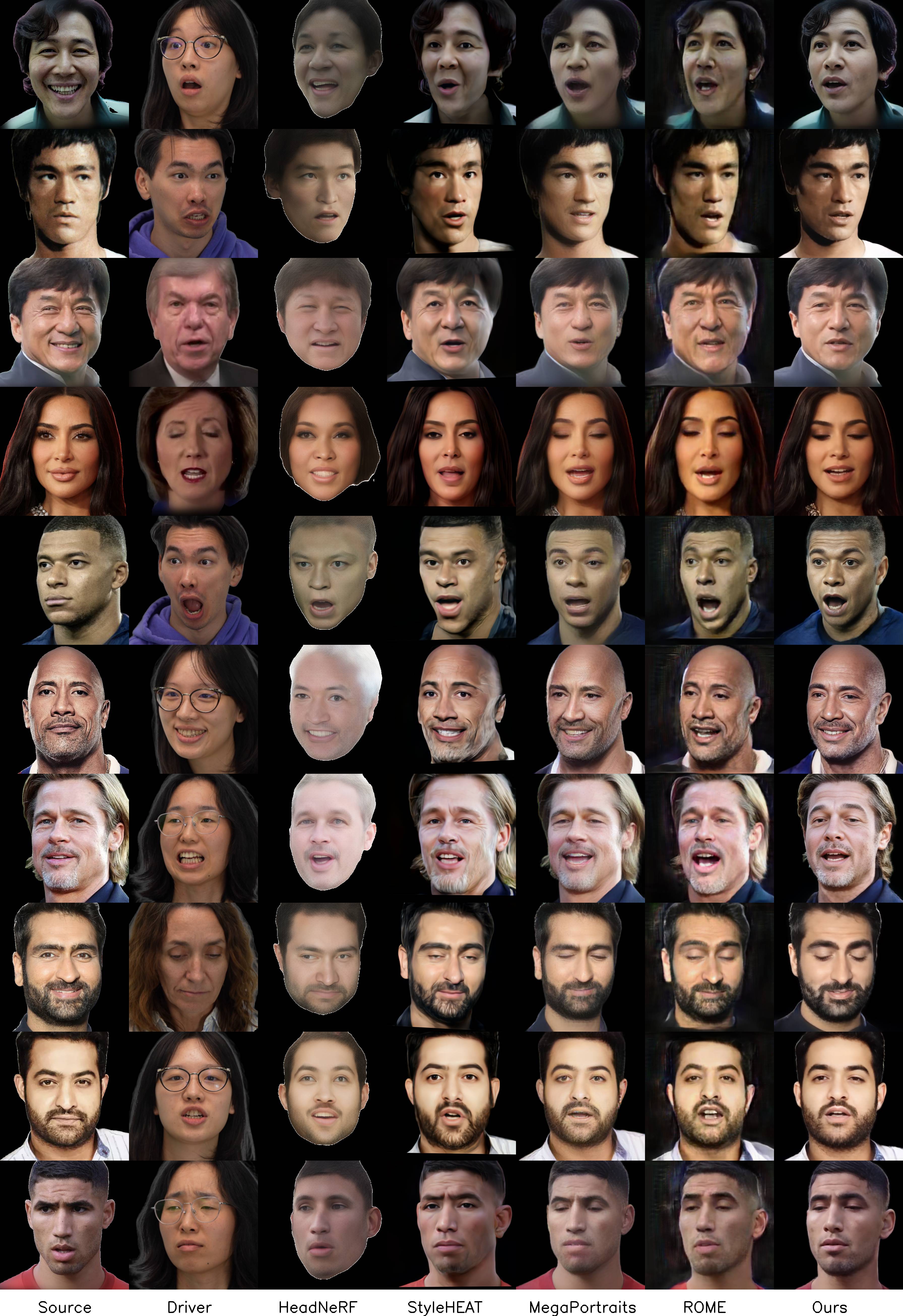}
    \caption{Qualitative results on various datasets.}
    \label{fig:supqal12}
\end{figure*}

\begin{figure*}
    \centering
    \includegraphics[width=0.85\linewidth]{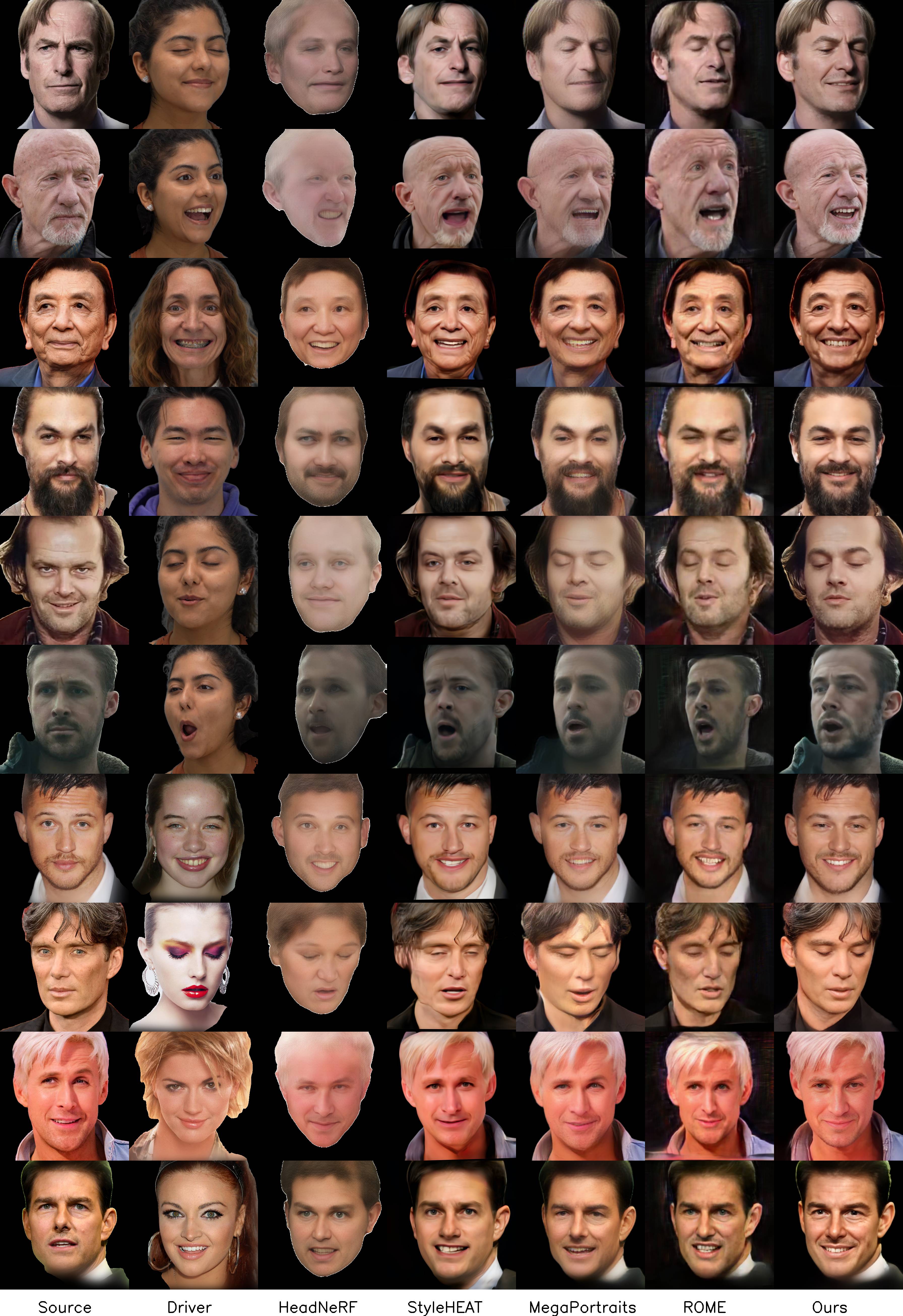}
    \caption{Qualitative results on various datasets.}
    \label{fig:supqal13}
\end{figure*}

\begin{figure*}
    \centering
    \includegraphics[width=0.85\linewidth]{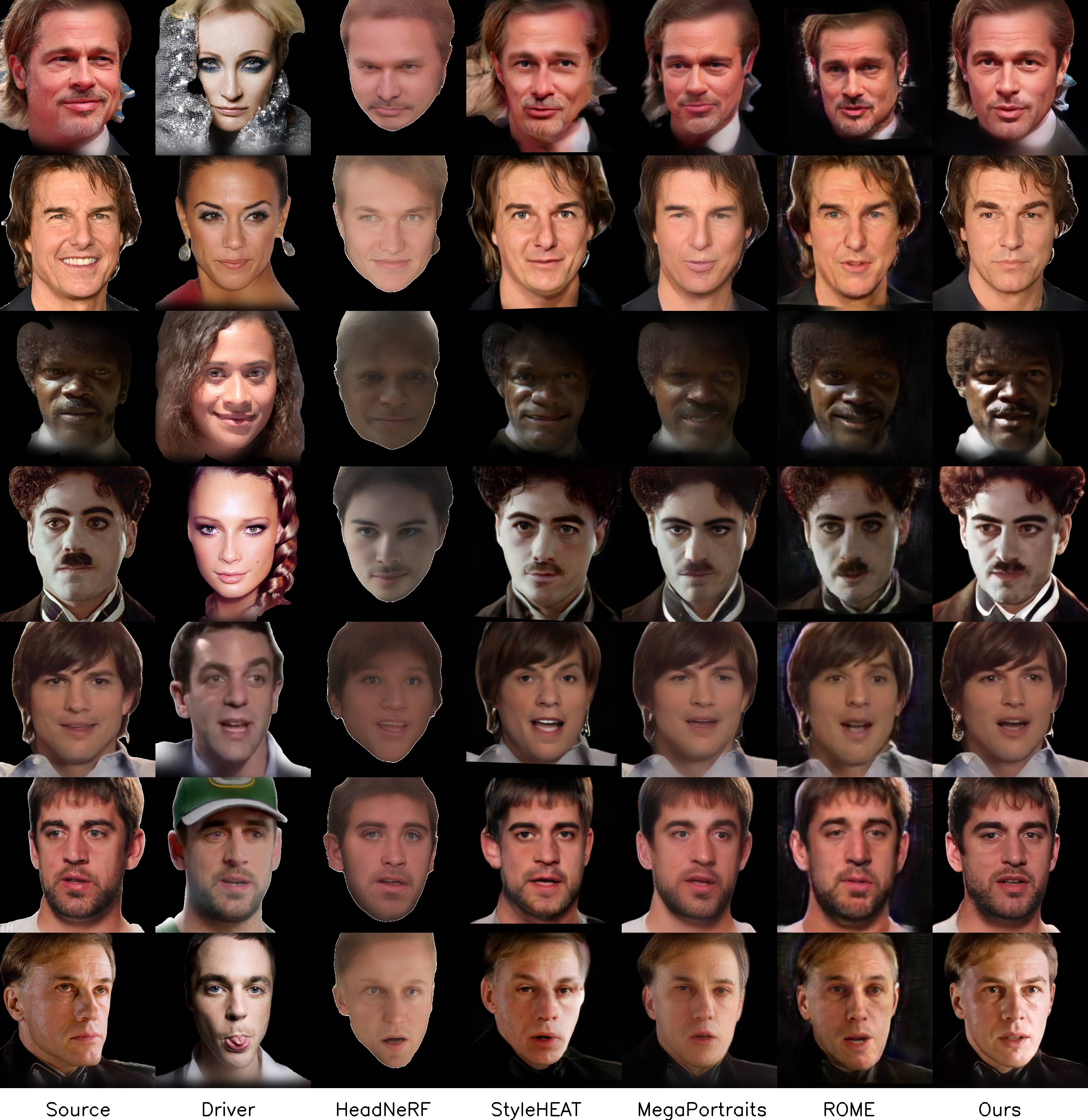}
    \caption{Qualitative results on various datasets.}
    \label{fig:supqual14}
\end{figure*}

We provide additional qualitative comparisons with other methods in \cref{fig:supqal01}, \cref{fig:supqal02}, \cref{fig:supqal03}, \cref{fig:supqal04}, \cref{fig:supqal05}, \cref{fig:supqal06}, \cref{fig:supqal07}, \cref{fig:supqal08}, \cref{fig:supqal09}, \cref{fig:supqal10}, \cref{fig:supqal11}, \cref{fig:supqal12}, \cref{fig:supqal13}, and \cref{fig:supqual14}.

\begin{figure*}[ht]
    \centering
    \includegraphics[width=0.85\linewidth]{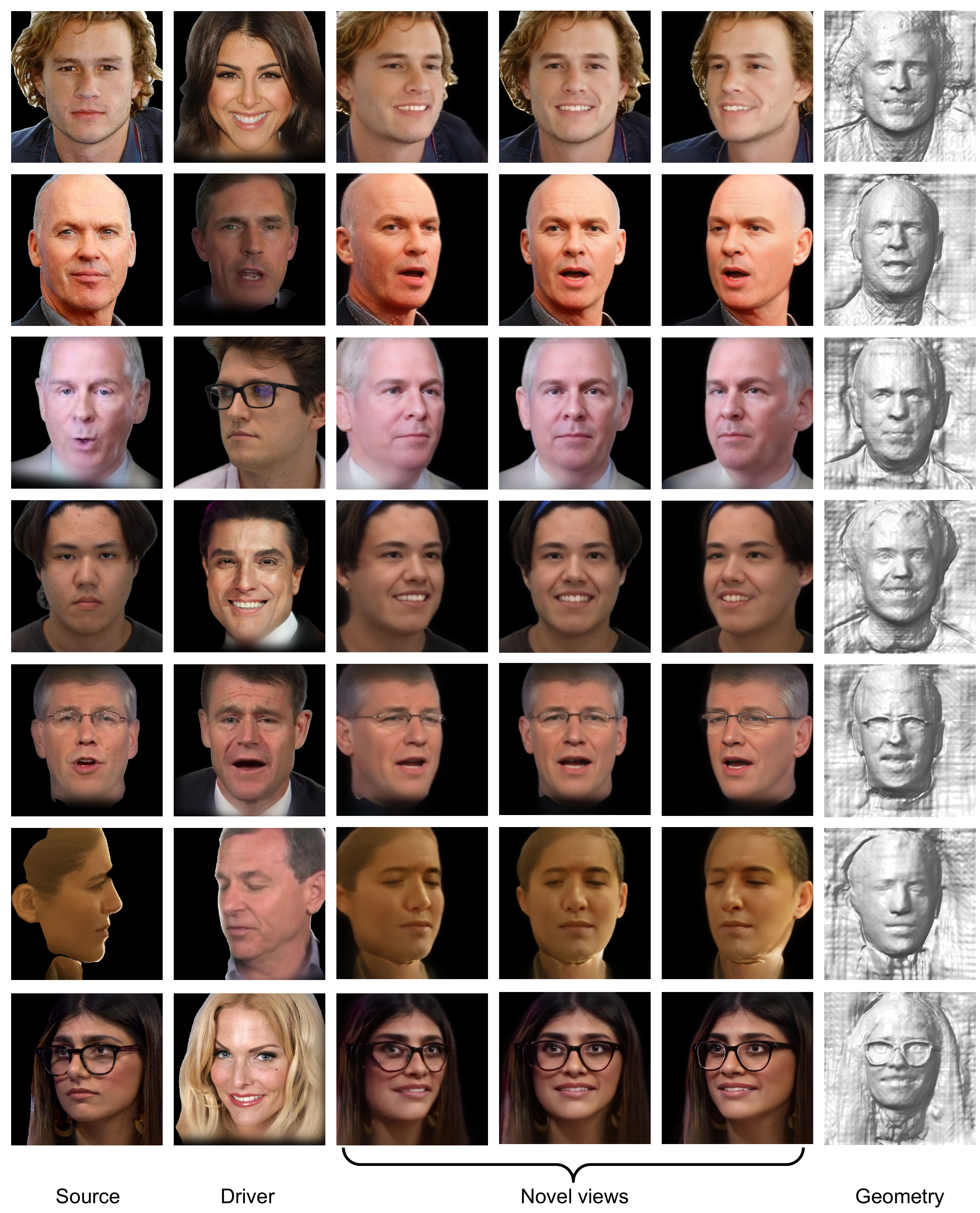}
    \caption{Synthesizing novel views using our method.}
    \label{fig:oursgeo}
\end{figure*}

In \cref{fig:oursgeo}, we evaluate the ability to synthesize novel views of our method. In addition, we also reconstruct the 3D mesh of the reenacted results.

\begin{figure*}
    \centering
    \includegraphics[width=0.95\linewidth]{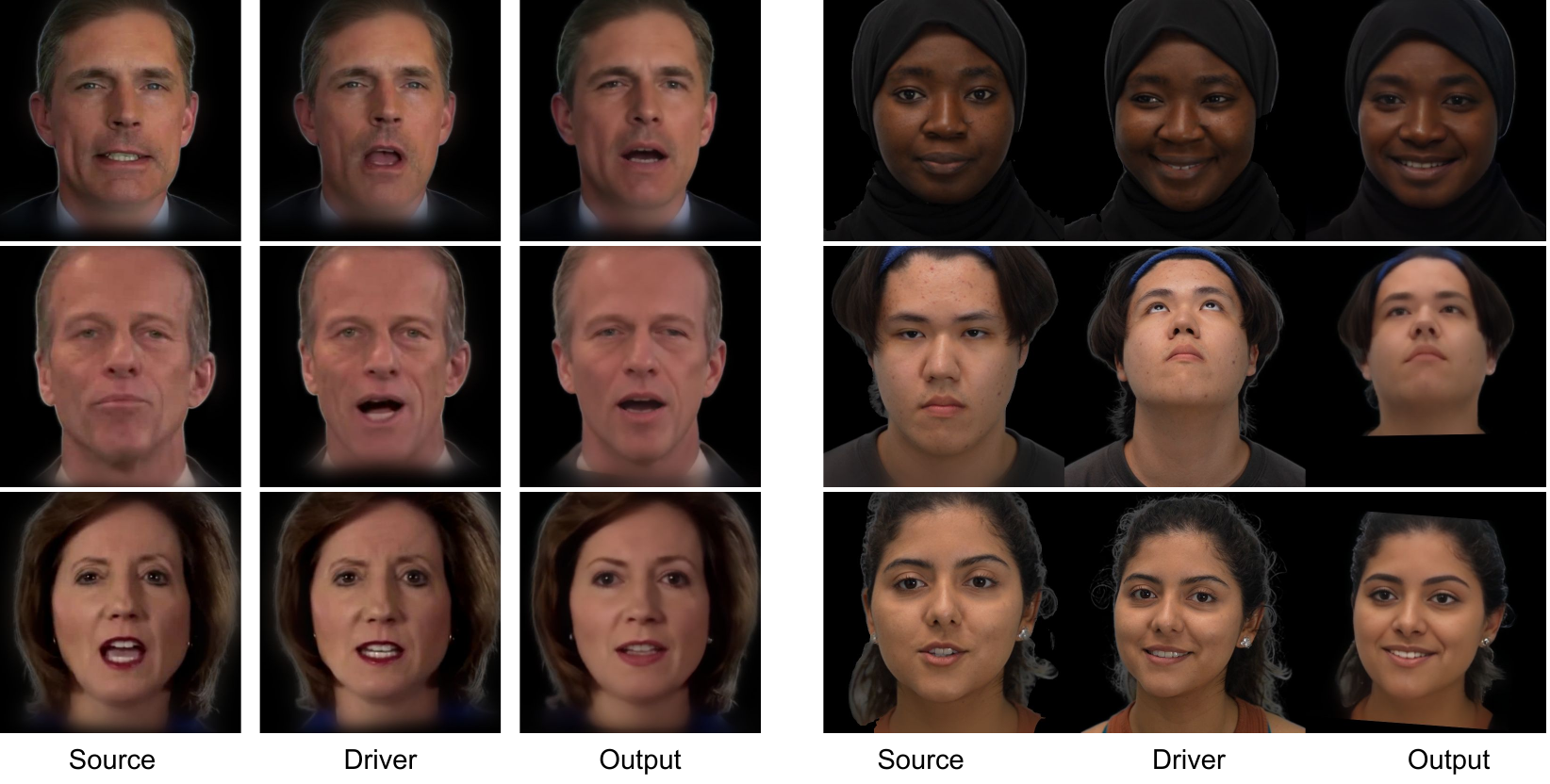}
    \caption{Qualitative results of our method on self-reenactment}
    \label{fig:self}
\end{figure*}

In \cref{fig:self}, we evaluate our model on self-reeactment task using HDTF and our collected datasets.

\begin{figure*}
    \centering
    \includegraphics[width=0.95\linewidth]{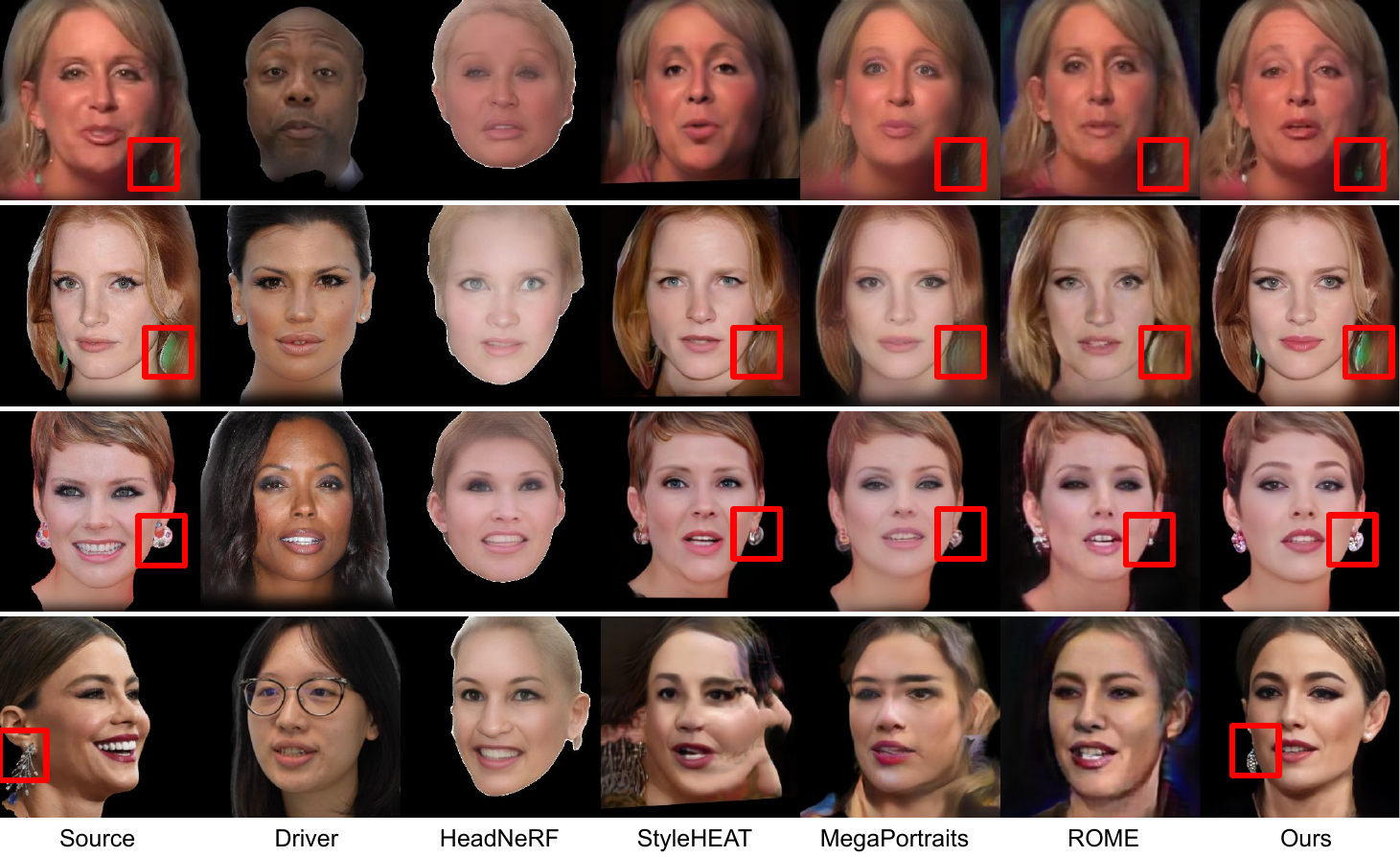}
    \caption{Our method faithfully retains the jewelries from the source image}
    \label{fig:jewelries}
\end{figure*}

In \cref{fig:jewelries}, we compares our method with the others on source images that have jewelries. As can be seen, other methods struggle to reconstruct the jewelries while our results still have the jewelries from the source input.

\section{Addtional Experiments with PTI \cite{roich2021pivotal}}
Our method can achieve high-quality results without noticeable identity change without additional fine-tuning, which is known to be computaionally expensive. In this section, we try to fine-tune \cite{roich2021pivotal} the super-resolution module using PTI \cite{roich2021pivotal} for 100 iterations, which takes around 1 minute per subject. Without PTI, our pipeline runs instantly similarly to \cite{li2023generalizable}. For most cases, the difference between results with and without fine-tuning is negligible. However, for out-of-domain images such as Mona Lisa, PTI fine-tuning helps retain the oil-painting style and fine-scale details from the input source. For the fine-tuning results, please refer to the supplementary video.

\section{Additional Limitations}
\begin{figure}
    \centering
    \includegraphics[width=0.9\linewidth]{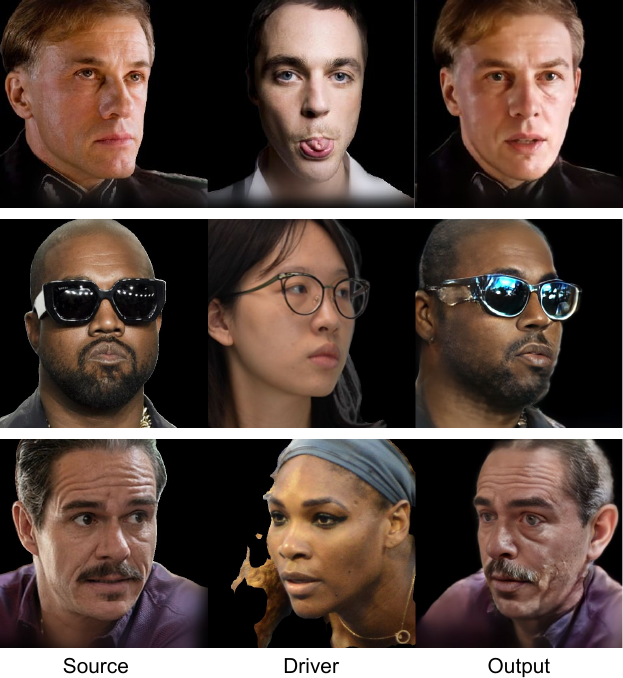}
    \caption{Additional Limitations: our method cannot handle the driver's tongue and sometimes produces wrong accessories that are out-of-domain, such as exotic sunglasses. Also, our head pose uses a single rigid transformation instead of a multi-joint body rig, which leads to the shoulders always moving together with the head pose.}
    \label{fig:limitationsext}
\end{figure}

Besides the limitations that we discussed in the paper, we also notice that the model cannot transfer tongue-related expressions or certain asymmetric expressions due to limited training data for our 3D lifting and expressions module. Since our method is not designed to handle the shoulder pose, the model uses the head pose as a single rigid transformation for the whole portrait. This issue would be an interesting research direction for future work. Also, our model sometimes fails to produce correct accessories when the input has out-of-distribution sunglasses. These failure cases are illustrated in \cref{fig:limitationsext}. 

\begin{figure}
    \centering
    \includegraphics[width=0.95\linewidth]{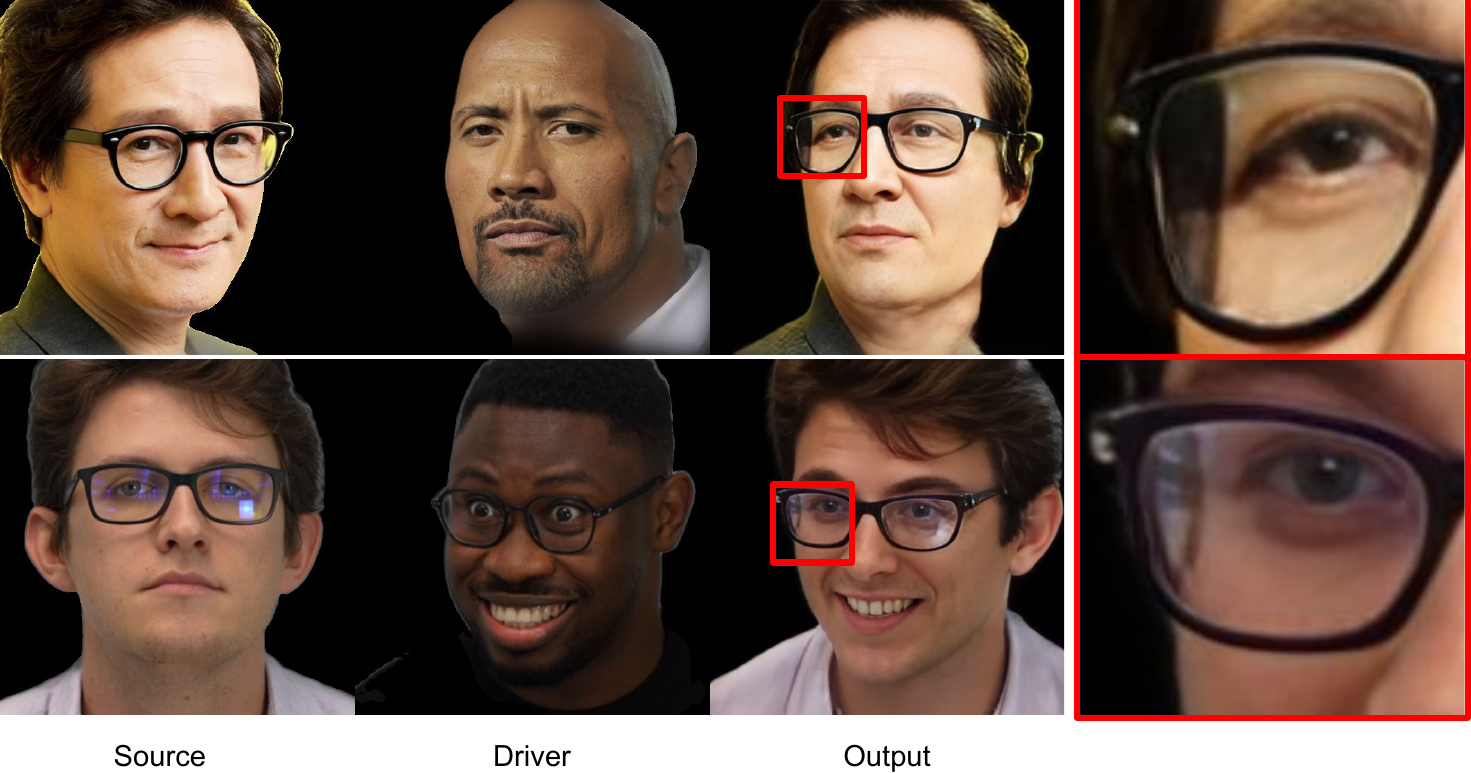}
    \caption{Our method can handle glass's refraction}
    \label{fig:refraction}
\end{figure}

{
    \small
    \bibliographystyle{ieeenat_fullname}
    \bibliography{main}
}


\end{document}